\documentclass{article}
\usepackage{sty,times}

\usepackage{amsmath,amsfonts,bm}

\def\eqref#1{equation~\ref{#1}}

\def\1{\bm{1}}

\DeclareMathAlphabet{\mathsfit}{\encodingdefault}{\sfdefault}{m}{sl}
\SetMathAlphabet{\mathsfit}{bold}{\encodingdefault}{\sfdefault}{bx}{n}

\def\gN{{\mathcal{N}}}

\newcommand{\E}{\mathbb{E}}

\usepackage{hyperref}
\usepackage{url}
\usepackage{algorithm}
\usepackage[endLComment=,italicComments=false]{algpseudocodex}
\usepackage{graphicx}
\usepackage{booktabs}
\usepackage{threeparttable}
\usepackage{longtable}
\usepackage{subcaption}
\usepackage{rotating}
\usepackage{enumitem}

\title{Unleashing Flow Policies with Distributional Critics}

\author{
  Deshu Chen\textsuperscript{1}\thanks{Email: \texttt{22110980001@m.fudan.edu.cn}} \quad
  Yuchen Liu\textsuperscript{1} \quad
  Zhijian Zhou\textsuperscript{1,2} \quad
  Chao Qu\textsuperscript{3}\thanks{Corresponding authors.} \quad
  Yuan Qi\textsuperscript{1}\footnotemark[2]   \bigskip\\
  \textsuperscript{1}Fudan University \\
  \textsuperscript{2}Shanghai Innovation Instutite \\
  \textsuperscript{3}INFLY TECH (Shanghai) Co., Ltd.
}

\begin{document}

\maketitle

\vspace{-2pt}
\begin{abstract}
\vspace{-2pt}

Flow-based policies have recently emerged as a powerful tool in offline and offline-to-online reinforcement learning, capable of modeling the complex, multimodal behaviors found in pre-collected datasets. However, the full potential of these expressive actors is often bottlenecked by their critics, which typically learn a single, scalar estimate of the expected return. To address this limitation, we introduce the Distributional Flow Critic (DFC), a novel critic architecture that learns the complete state-action return distribution. Instead of regressing to a single value, DFC employs flow matching to model the distribution of return as a continuous, flexible transformation from a simple base distribution to the complex target distribution of returns. By doing so, DFC provides the expressive flow-based policy with a rich, distributional Bellman target, which  offers a more stable and informative learning signal. Extensive experiments across D4RL and OGBench benchmarks demonstrate that our approach achieves strong performance, especially on tasks requiring multimodal action distributions, and excels in both offline and offline-to-online fine-tuning compared to existing methods.
\end{abstract}
\vspace{-3pt}
\section{Introduction}

In modern reinforcement learning, particularly in offline and offline-to-online settings, a central challenge is learning effective policies from complex, pre-collected datasets \citep{fujimoto2021minimalist,tarasov2023corl,parkogbench}. To this end, flow-based policies, trained with generative techniques like flow matching, represent a significant advance \citep{lipman2023flow, zhang2025energy}. Unlike traditional parameterizations, such as a unimodal Gaussian, flow-based methods can capture the rich, often multimodal action distributions inherent in diverse expert data. The success of such agents is measured not only by their static offline performance but also by their capacity for safe and efficient online fine-tuning.

Despite the success of these expressive policies, a critical limitation persists in their underlying learning mechanism. Current state-of-the-art flow-based methods pair a highly expressive actor (trained with flow matching) with a comparatively simple critic that only estimates the expected value of future returns, \(Q(s,a)\) \citep{dabney2018distributional}. This creates an ``information bottleneck": the critic provides a single scalar value as a learning signal, which is a low-dimensional summary of a potentially complex and stochastic future. This simplistic signal may be insufficient to effectively guide the training of a high-capacity, multimodal policy.  We argue that for an expressive policy to reach its full potential, it requires an equally expressive critic. The critic should provide a richer, more informative signal that captures not just the expected outcome, but the entire distribution of possible returns. This distributional perspective, pioneered by \citep{bellemare2017distributional,dabney2018implicit,morimura2010parametric}, is known to improve learning stability and performance.

To bridge this crucial gap, we introduce the \textbf{Distributional Flow Critic (DFC)}, a novel architecture designed to learn the entire return distribution, $Z(s,a)$, using flow matching. However, a naive application of flow matching to established distributional losses faces a significant hurdle: estimating the loss requires sampling from the learned distribution $Z_{\phi}(s,a)$, which necessitates backpropagating through the entire trajectory of an ODE solver. This process is known to be computationally expensive and notoriously unstable, a challenge also observed in flow-based policy learning \citep{park2025flow}. To circumvent this, we propose an elegant \textbf{distillation-based architecture} comprising two critic networks. A multi-step \textit{target flow critic} captures the complex target distribution, while a single-step  critic efficiently distills this distributional knowledge, ensuring stable and effective training. To ground our approach in a strong and relevant context, we integrate DFC into the Flow Q-learning framework \citep{park2025flow}, a state-of-the-art method known for its robust performance in offline and offline-to-online tasks. This allows our distributional critic to provide a rich, nuanced learning signal to an already powerful flow-based actor. Although our experiments focus on this specific integration, DFC is designed as a modular component. This suggests its potential as a drop-in replacement for standard scalar critics in other off-policy actor-critic algorithms, particularly those that also aim to capture complex, multimodal policies.

By creating a symbiotic architecture where both the actor and the critic leverage the power of flow matching, our method effectively addresses the dual challenges of behavioral complexity and value uncertainty. To validate our central claim, \textit{``A distributional critic unlocks superior performance for flow-based policies.''}, we benchmark DFC against a suite of state-of-the-art flow-policy methods. We conduct extensive evaluations on both pure offline and challenging offline-to-online  benchmarks. Our results show that our method not only excels in the static offline phase but also demonstrates superior adaptability and sample efficiency during online fine-tuning. This confirms that equipping expressive policies with an equally expressive distributional critic is a critical step forward for both offline pre-training and subsequent online adaptation.

Our main contributions are:

\textbf{A Distributional Flow-Based Critic.} We introduce a flow-matching model for the critic that moves beyond traditional scalar Q-value estimation. This model learns to generate the entire distribution of future returns, providing a richer, more stable, and more informative learning signal for policy optimization, which is especially beneficial in highly stochastic environments. We conduct an ablation study to demonstrate the necessity of our proposed design choices.

\textbf{State-of-the-Art Empirical Performance.} Through extensive experiments on a wide range of challenging benchmarks, we demonstrate that our method significantly outperforms existing approaches, particularly on tasks that require multimodal action distributions. We also show that our framework is highly effective for offline-to-online fine-tuning. We empirically show that our method consistently outperforms existing state-of-the-art approaches, registering a comprehensive performance gain of more than $10\%$. 

\vspace{-3pt}
\section{Background}

\subsection{Distributional Reinforcement Learning}
\vspace{-3pt}
We model the agent-environment interaction as a Markov Decision Process (MDP)~\citep{sutton2018mdp}, formally defined by the tuple $(\mathcal{S}, \mathcal{A}, P, r, \gamma)$. Here, $\mathcal{S}$ is the state space, $\mathcal{A}$ is the action space, $P: \mathcal{S} \times \mathcal{A} \to \Delta(\mathcal{S})$ is the transition probability function where $\Delta(\cdot)$ denotes the set of probability distributions over a space, $r: \mathcal{S} \times \mathcal{A} \to \mathbb{R}$ is the reward function, and $\gamma \in [0, 1)$ is the discount factor. An agent's behavior is described by a policy $\pi: \mathcal{S} \to \Delta(\mathcal{A})$, which maps states to a probability distribution over actions. The goal of the agent is to learn a policy that maximizes the expected discounted cumulative return: $J(\pi) = \mathbb{E}_{\tau \sim \pi} \left[ \sum_{t=0}^{\infty} \gamma^t r(s_t, a_t) \right],$
where the trajectory $\tau = (s_0, a_0, s_1, a_1, \dots)$ is generated by executing policy $\pi$ (i.e., $s_0 \sim p_0(s)$, $a_t \sim \pi(\cdot|s_t)$, and $s_{t+1} \sim P(\cdot|s_t, a_t)$). The action-value function  $Q^\pi(s, a)$ is the expected return after taking action $a$ in state $s$ and subsequently following $\pi$. The Q-function satisfies the Bellman expectation equation:
$
Q^\pi(s, a) = \mathbb{E}_{s' \sim P(\cdot|s,a)} \left[ r(s, a) + \gamma \mathbb{E}_{a' \sim \pi(\cdot|s')} [Q^\pi(s', a')] \right].
$

Standard reinforcement learning focuses on estimating the expected value of the cumulative return, $Q^\pi(s, a)$. Distributional Reinforcement Learning (DRL)~\citep{bellemare2017distributional} extends this paradigm by learning the entire probability distribution of the return. The return, or sum of discounted future rewards, is treated as a random variable, denoted by $Z^\pi(s, a)$:
\begin{equation}\label{equ:distributional_bellman_equation}
Z^\pi(s, a) = \sum_{t=0}^{\infty} \gamma^t r(S_t, A_t), \quad \text{where } S_0=s, A_0=a.
\end{equation}
The core of DRL is the \emph{distributional Bellman equation}, which states that the distribution of the return at the current state-action pair is related to the distribution of the return at the next state: $\mathcal{T}^{\pi} Z(s, a) \stackrel{D}{:=} r(s, a) + \gamma Z^{\pi}(S', A'),$
where $\stackrel{D}{=}$ denotes equality in distribution, $S' \sim \mathcal{P}(\cdot|s, a)$, and $A' \sim \pi(\cdot|S')$.

In practice, DRL algorithms learn a parameterized approximation $Z_{\theta}(s,a)$ of the true return distribution. For example, C51~\citep{bellemare2017distributional} models the distribution with a discrete set of fixed supports (atoms) and learns the corresponding probabilities. Quantile Regression DQN (QR-DQN)~\citep{dabney2018distributional} takes a different approach by directly learning the quantile function of the return distribution.

A natural metric for comparing return distributions is the \textbf{Wasserstein distance}~\citep{villani2003topics}. Minimizing the distributional Bellman error with respect to the 1-Wasserstein distance, $W_1(\mathcal{T}^{\pi} Z_{\theta}(s,a) , Z_{\theta}(s,a))$, is a primary objective. For one-dimensional distributions $\mu_X$ and $\mu_Y$ with cumulative distribution functions (CDFs) $F_X$ and $F_Y$, this distance has a closed-form solution: $W_1(\mu_X, \mu_Y) = \int_0^1 |F_X^{-1}(\tau) - F_Y^{-1}(\tau)| d\tau$.
Given $N$ i.i.d. samples $\{x_i\}_{i=1}^N$ from $\mu_X$ and $\{y_i\}_{i=1}^N$ from $\mu_Y$, the distance can be approximated by sorting the samples (denoted by $\tilde{x}_i, \tilde{y}_i$): $\hat{W}_1(\mu_X, \mu_Y) \approx \frac{1}{N} \sum_{i=1}^N |\tilde{x}_i - \tilde{y}_i|.$ However, this sample-based approximation is biased. QR-DQN elegantly circumvents this issue. Instead of directly minimizing an approximated Wasserstein distance, it minimizes the \textbf{quantile regression loss} ~\citep{koenker1978regression}. This objective implicitly minimizes the Wasserstein distance between the empirical distributions of samples, providing a more stable and effective method for learning the return distribution.

\subsection{Flow Matching and Flow Q-learning}

Flow Matching is a technique for training expressive generative models by learning a vector field $v_\theta$ that defines a mapping from a simple noise distribution to a complex target distribution by solving an ordinary differential equation (ODE) ~\citep{lipman2023flow}. This framework can be used to create highly expressive \textit{flow policies}, $\pi_\theta(a|s)$, capable of representing complex, multimodal action distributions, which is a significant advantage over simpler parametric forms like Gaussians. However, directly training these policies with reinforcement learning objectives is challenging, as it requires backpropagating gradients through the ODE solver, a process that is computationally expensive and notoriously unstable.

Flow Q-Learning (FQL)~\citep{park2025flow} is an offline and offline-to-online reinforcement learning method that elegantly solves this problem by decoupling the policy's training from direct value maximization. The core idea is to first train an iterative flow policy using only a behavioral cloning (BC) objective on the offline dataset. This BC flow policy, denoted $\mu_\theta(s, \epsilon)$ where $\epsilon$ is the initial noise, learns to capture the complex action distribution of the dataset. While \(\mu_\theta\) is deterministic function, the random sampling of \(\epsilon\) allows it to function as a stochastic policy, formally expressed as \(\pi_\theta(a \mid s)\). In this paper, we slightly abuse notation for simplicity and refer to both \(\mu_\theta\) and \(\pi_\theta\) as the ``policy".

Instead of directly optimizing this iterative policy with RL, FQL introduces a separate, expressive one-step policy, $\pi_\omega$, which is trained to maximize Q-values while being regularized via distillation from the BC flow policy. This distillation loss prevents the one-step policy from deviating too far from the learned data distribution. The distillation loss is defined as the mean squared error between the two policies' outputs for a given state and noise vector:
\begin{equation}\label{equ:actor_loss}
\mathcal{L}_{\text{Distill}}(\omega) = \mathbb{E}_{s \sim \mathcal{D}, \epsilon \sim \mathcal{N}(0, I)} \left[ \| \pi_{\omega}(s, \epsilon) - \mu_{\theta}(s, \epsilon) \|^2 \right]
\end{equation}
The complete objective for the one-step policy $\pi_\omega$ combines the Q-maximization term with this distillation loss, which acts as a behavioral cloning regularizer:
\begin{equation}\label{equ:FQL_objective}
\mathcal{L}_{\pi}(\omega) = \mathbb{E}_{s \sim \mathcal{D}, a^{\pi} \sim \pi_{\omega}} [-Q_{\phi}(s, a^{\pi})] + \alpha \mathcal{L}_{\text{Distill}}(\omega)
\end{equation}
This approach allows FQL to leverage the expressivity of the flow model while using a stable, one-step update for the RL objective, thereby avoiding unstable backpropagation through the ODE solver.

\begin{figure}[htbp]
    \centering
    \includegraphics[width=\textwidth]{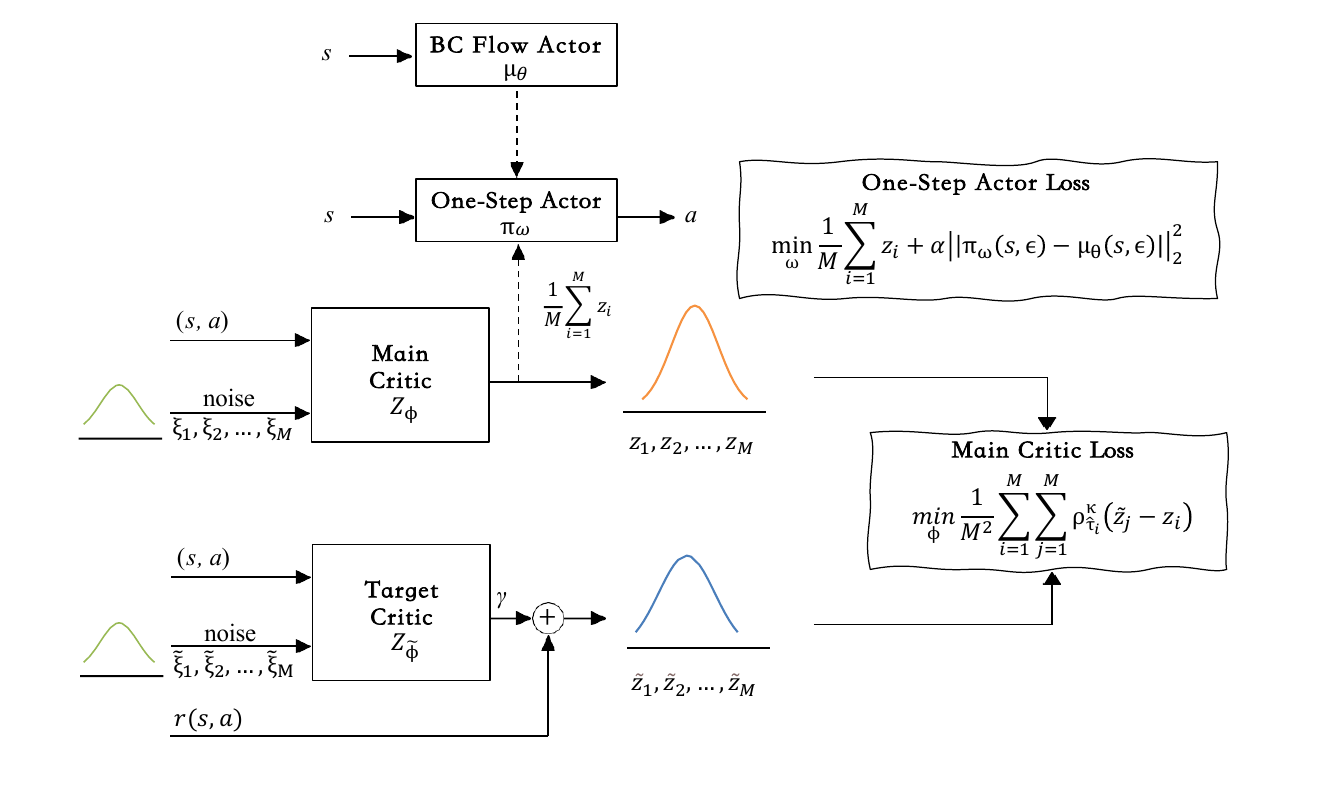}
    \caption{\textbf{A schematic of the DFC architecture.} The illustration delineates the decoupled training paradigm, wherein the target distributional flow critic, $Z_{\tilde{\phi}}$, is updated to form a set of target return samples, which in turn supervises the main distributional critic, $Z_\phi$. The synergistic output of both critics is then utilized to optimize the dual-policy actor.}
    \label{fig:dfc_diagram}
    \vspace{-10pt}
\end{figure}

\section{Methodology}

Our core contribution is a novel distributional critic based on flow matching, designed to enhance the expressive power of any actor-critic method that employs a flow-based policy. In principle, our critic can be integrated into various such frameworks. To demonstrate its effectiveness, we build upon the state-of-the-art Flow Q-Learning (FQL) framework, chosen for its proven stability and performance.

\vspace{-1pt}
\subsection{Actor Architecture}
\vspace{-1pt}
For our actor, we directly adopt the dual-policy architecture from Flow Q-Learning (FQL). As previously described, this framework uses a one-step policy, $\pi_\omega$, which is updated to maximize Q-values while being regularized via distillation from an expressive, behaviorally-cloned flow policy, $\mu_\theta$. The objective function is similar to Equation \ref{equ:FQL_objective}, and will elaborate how to estimate it after we define the critic later.
While we retain this actor structure, our core contribution lies in the novel formulation of the critic function $Q_{\phi}(s, a^{\pi})$, which is designed to match the actor's expressive power and is detailed next.

\subsection{Critic as a Conditional Distributional Flow Model}

A critical limitation in actor-critic methods arises when an expressive actor is paired with a simple critic. A standard critic that predicts only the mean of the return cannot capture the potentially complex value distributions---such as multimodal or skewed returns---that a powerful actor might induce. This mismatch can lead to an impoverished learning signal and unstable training.

To address this, we propose a distributional critic that learns the entire distribution of returns. A naive approach would be to model the critic $Z_{\phi}(s,a)$ as a single conditional flow model trained to minimize the Wasserstein distance to the target distribution, e.g., $W_1(\mathcal{T}^{\pi} Z_{\phi}(s,a), Z_{\phi}(s,a))$. However, this approach is infeasible for two key reasons: \\
\textbf{Unstable Critic Training.} Generating samples from $Z_{\phi}$ to compute the loss would require solving an ODE. Training would thus necessitate backpropagation through the ODE solver's steps, which is notoriously unstable. \\
\textbf{Unstable Actor Training.} The actor update requires the gradient $\nabla_a Q(s,a)$. If the Q-value is the mean of the flow-generated distribution $Z_{\phi}$, computing this gradient would also require backpropagation through the ODE solver, which the very problem FQL was designed to avoid.

To circumvent this issue, we introduce a two-stage architecture comprising a \textbf{target distributional flow critic}, $Z_{\tilde{\phi}}$, and a \textbf{main distributional critic}, $Z_\phi$. This approach is analogous to knowledge distillation: the powerful, multi-step flow model $Z_{\tilde{\phi}}$ learns the complex target distribution, and its knowledge is then distilled into the simpler, one-step critic $Z_\phi$, which can be updated efficiently.

\textbf{Learning the Target Distribution with Flow Matching.} The target critic $Z_{\tilde{\phi}}$ is a conditional flow model designed to approximate the target value distribution $\mathcal{T}^{\pi}Z(s,a)$. Its velocity field, $v_{\tilde{\phi}}$, is updated using the flow matching objective. Specifically, for a given transition $(s, a, r, s')$, we first construct a set of target return samples $\{\tilde{z}_j\}_{j=1}^{M}$ using the distributional Bellman equation:
\begin{equation}
    \tilde{z}_j = r + \gamma z'_j, \quad \text{where} \quad z'_j =  Z_{\tilde{\phi}}(s', \pi_\omega(s'), \tilde{\xi}_j).
\end{equation}
Here, samples $z'_j$ are generated by solving the ODE defined by the target critic $Z_{\tilde{\phi}}$ itself (with frozen parameters from a previous iteration, akin to a standard target network). Then, the parameters $\tilde{\phi}$ are updated by minimizing the flow matching loss, which trains the model to transport samples from a base Gaussian distribution to this target distribution:
\begin{equation}
    \mathcal{L}_{\text{Flow}}(\tilde{\phi}) = \mathbb{E}_{\tau, \tilde{\xi}, \tilde{z}}\left[ ||{v_{\tilde{\phi}}(\tau, (1-\tau)\tilde{\xi}+\tau\tilde{z} | s, a) - (\tilde{z} - \tilde{\xi})}||_2^2 \right],
\end{equation}
where $\tau \sim \mathcal{U}(0,1)$, $\tilde{\xi} \sim \mathcal{N}(0, I_d)$, and $\tilde{z}$ is a sample from the target set constructed above.

\textbf{Distilling the Target Distribution via Quantile Regression.} After updating the target flow network, we train the main critic $Z_\phi$. The purpose of $Z_\phi$ is to learn a direct, one-step mapping to a distribution that matches the output of the target flow critic $Z_{\tilde{\phi}}$. This avoids the need for an ODE solver during the actor update. We achieve this distillation by minimizing the 1-Wasserstein distance between the two distributions, using the quantile Huber loss as a practical and robust proxy ~\citep{dabney2018distributional}.

We draw a set of $M$ samples $\{\tilde{z}_j\}$ from the target critic $Z_{\tilde{\phi}}$ and another set of $M$ quantile values $\{z_i\}$ from the main critic $Z_\phi$. The main critic's parameters $\phi$ are then updated by minimizing the following loss:
\begin{equation}
    \mathcal{L}_{\text{critic}}(\phi) = \frac{1}{M^2} \sum_{i=1}^{M} \sum_{j=1}^{M} \rho_{\hat{\tau}_i}^{\kappa} (\tilde{z}_{j} - z_{i} ),
\end{equation}
where $\rho_{\hat{\tau}}^{\kappa}$ is the quantile Huber loss function defined as:
\begin{equation}
    \rho^{\kappa}_{\hat{\tau}} (u) = |\hat{\tau} - \mathbb{I}(u<0) | \mathcal{L}_\kappa(u), \quad \text{with} \quad \mathcal{L}_\kappa(u) =
    \begin{cases}
        0.5 u^2 & \text{if } |u| \le \kappa \\
        \kappa(|u| - 0.5 \kappa) & \text{otherwise.}
    \end{cases}
\end{equation}
The quantile levels are set to $\hat{\tau}_i = (2i-1)/(2M)$ for $i=1, \dots, M$. This loss effectively trains $Z_\phi$ to produce a distribution that mirrors the one generated by the more complex target flow critic $Z_{\tilde{\phi}}$. The actor is then updated using the distilled critic $Z_\phi$, sidestepping the BPTT problem entirely.

\subsection{Actor Update}

With our trained main critic $Z_\phi$, we can now update the actor policy $\pi_\omega$. Following common practice in distributional RL \citep{barth2018distributed}, we estimate the Q-value as the mean of the return distribution. 
The actor $\pi_\omega$ is then trained to produce actions that maximize this estimated Q-value. The final objective for the actor combines the Q-maximization term with the FQL distillation regularizer:
$$\E[-\frac1M\sum_{i=1}^{M}[Z_\phi(s, \mu_{\omega}(s, \epsilon), \xi_i)] + \alpha \|\mu_{\omega}(s, \epsilon) - \mu_\theta(s, \epsilon)\|_2^2].$$
Because $Z_\phi$ is a simple feed-forward network, computing the policy gradient is stable and efficient, entirely sidestepping the BPTT problem.

All the steps of our algorithm are described in Algorithm \ref{alg:1}. The synergistic integration of our distributional critic and dual-policy actor forms a holistic framework with significant advantages. The flow-based critic architecture allows us to model the entire value distribution with unprecedented fidelity, capturing nuances such as multimodality and skewness that are invisible to traditional, expectation-based critics. This rich distributional information provides a more robust and informative gradient signal to the actor, moving beyond a simple scalar reward signal. The design not only enhances training stability but also allows each component to leverage its respective strengths: the expressiveness of flow models for target generation and the efficiency of quantile regression for distributional matching. The result is a robust framework for accurately approximating complex value distributions in reinforcement learning.

\begin{algorithm*}[htbp]
\caption{Distributional Flow Critic (DFC)}
\label{alg:1}
\begin{algorithmic}

\For{the number of environment steps}

\State Sample batch $\{(s, a, r, s')\}$

\LComment{Train vector field $v_{\tilde{\phi}}$ in distributional flow critic $Z_{\tilde{\phi}}$}
\State Sample noise $\tilde{\xi}_1, \tilde{\xi}_2, \dots, \tilde{\xi}_{M} \sim \gN(0, I_d)$
\State \( \tau \sim \mathcal{U}(0,1)\)
\For{$j =1, \dots, M$}
\State $\tilde{z}_j \gets r + \gamma Z_{\tilde{\phi}}(s', \pi_\omega(s'), \tilde{\xi}_j)$ \Comment{Bellman Update}
\State \(\tilde{z}^\tau_j \gets (1-\tau)\tilde{\xi}_j+\tau\tilde{z}_j\)
\EndFor
\State Update {$\tilde{\phi}$} to minimize $\E\left[\frac{1}{M}\sum_{j=1}^{M}\|v_{\tilde{\phi}}(\tau, s, \tilde{z}_j^\tau) - (\tilde{z}_j - \tilde{\xi}_j)\|_2^2\right]$ 
\LComment{Train distributional critic $Z_\phi$}
\State Sample noise $\xi_0, \xi_1, \dots, \xi_{M-1} \sim \gN(0, I_d)$, 
\State $z_j \gets Z_\phi(s,a,\xi_j)$ for $j =1, \dots, M$
\State Update $\phi$ to minimize $\E\left[\frac{1}{M^2} \sum_{i=1}^{M} \sum_{j=1}^{M} \rho_{\hat{\tau}_i}^{\zeta} (\tilde{z}_{j} - z_{i}) \right]$
\LComment{Train vector field $v_\theta$ in flow policy $\pi_\theta$}
\State Sample noise $x^0 \sim \gN(0, I_d)$
\State $x^1 \gets a$
\State Sample $t \sim \mathcal{U}([0, 1])$
\State $x^t \gets (1-t) x^0 + tx^1$
\State Update {$\theta$} to minimize $\E[\|v_{\theta}(t, s, x^t) - (x^1 - x^0)\|_2^2]$
\LComment{Train one-step policy $\pi_\omega$}
\State Sample noise $\epsilon \sim \gN(0, I_d)$
\State Update {$\omega$} to minimize $\E[-\frac1M\sum_{i=1}^{M}[Z_\phi(s, \mu_{\omega}(s, \epsilon), \xi_i)] + \alpha \|\mu_{\omega}(s, \epsilon) - \mu_\theta(s, \epsilon)\|_2^2]$
\EndFor
\Return One-step policy $\pi_\omega$

\end{algorithmic}
\end{algorithm*}

\section{Related Works}

\subsection{Diffusion and Flow Policy RL}

Recent advancements in iterative generative modeling, particularly denoising diffusion~\citep{sohl2015deep, ho2020denoising} and flow matching~\citep{lipman2023flow, esser2024scaling}, have spurred significant innovation within reinforcement learning (RL) due to their capacity to model complex, high-dimensional data distributions.

Previous works can be broadly categorized based on how they leverage these generative models. One line of research applies them to high-level decision-making tasks such as planning and hierarchical learning \citep{janner2022planning, ajay2023is, zheng2023guided, liang2023adaptdiffuser, suh2023skill, venkatraman2024learning, chen2024hierarchical}. Another prominent application is to use generative models create synthetic environments or supplement training data to improve policy robustness and generalization \citep{lu2023synther, ding2024diffusion, jackson2024policy, alonso2024learning}. A third area involves the use of generative models to improve strategy implementation \citep{mazoure2020leveraging, ren2024diffusion} or to model policies \citep{park2025flow}. 

Several strategies have been proposed to train flow and diffusion policies for offline reinforcement learning. These can be broadly categorized into three main paradigms: 1) \textbf{Value-weighted regression}, which prioritizes high-advantage actions but can be limited by the expressivity of policies trained on static datasets~\citep{lu2023qgpo, kang2023edp, zhang2025energy}; 2) \textbf{Reparameterization-based policy gradients}, which optimize the value function by backpropagating through the policy but often suffer from instability and high variance~\citep{wang2023dql, he2023diffcps, ding2024consistencyac}; and 3) \textbf{Rejection sampling}, which enhances stability by filtering actions but incurs substantial computational overhead~\citep{chen2023sfbc, hansen-estruch2023idql, he2024aligniql}. Beyond these, other techniques include direct action gradients~\citep{yang2023dipo, psenka2024qsm, li2024ddiffpg}, bi-level MDP formulations~\citep{ren2024diffusion}, and integrations with implicit Q-learning~\citep{chen2024srpo, chen2024dtql}.

In this work, we employ a reparameterized policy gradient approach. Distinct from prior methods such as Diffusion-QL~\citep{wang2023dql}, DiffCPS~\citep{he2023diffcps}, Consistency-AC~\citep{ding2024consistencyac}, SRDP~\citep{ada2024srdp}, and EQL~\citep{zhang2024entropydql}, our method circumvents backpropagation through the generative model by instead training a one-step policy alongside a distributional critic. This strategy notably improves training stability and computational efficiency. Empirically, we demonstrate that our approach achieves a significant performance leap over existing baselines.

\subsection{Distributional RL}

Distributional Reinforcement Learning (DRL) marked a paradigm shift from modeling expected returns to capturing the full distribution of stochastic outcomes~\citep{bellemare2017distributional}. Early value-based methods for discrete action spaces, such as C51~\citep{bellemare2017distributional}, parameterized the return distribution with discrete atoms. This was advanced by QR-DQN~\citep{dabney2018distributional} and IQN~\citep{dabney2018implicit}, which respectively used discrete quantiles and learned a full continuous quantile function for enhanced flexibility. The extension of DRL to continuous control yielded policy-gradient methods like D4PG~\citep{barth-maron2018distributed}, which demonstrated strong performance but inherited the representational limitations of a discrete categorical critic. To overcome this constraint, which makes it difficult to capture complex return distributions, SDPG~\citep{singh2022sample-based} introduced a sample-based distributional policy gradient method that models the return distribution via reparameterization, thus avoiding the constraints of discrete representations.

Building upon this trajectory, our work further leverages flow matching as the core generative mechanism for modeling distributions in RL. Unlike previous sample-based methods that often rely on adversarial training, our approach benefits from the unique advantages inherent to flow matching. Firstly, flow matching offers superior expressiveness by directly learning a continuous-time vector field that transforms a simple noise distribution into any arbitrarily complex target distribution. By integrating this powerful generative tool into both the critic and actor, our method provides a robust and highly flexible framework for distributional reinforcement learning.

\section{Experiment}

\subsection{Benchmarks}

Our empirical evaluation is conducted on two comprehensive benchmark suites: the widely-adopted D4RL benchmark~\citep{fu2020d4rl} and the recently proposed OGBench~\citep{park2025ogbench}, which together provide a diverse set of challenges spanning locomotion, manipulation, and both state-based and pixel-based observations.

\textbf{D4RL.} From the D4RL benchmark, we select a suite of particularly challenging tasks to rigorously test our algorithm's capabilities. Specifically, we evaluate on six distinct \texttt{antmaze} navigation tasks and twelve \texttt{adroit} manipulation tasks. For \texttt{adroit} tasks, performance is reported using the standard normalized return score~\citep{fu2020d4rl}, calculated as: $\text{Normalized Return} = 100 \times \frac{\text{score} - \text{random\_score}}{\text{expert\_score} - \text{random\_score}}$.
For \texttt{antmaze} tasks, we report the success rate, which measures the percentage of episodes where the agent successfully completes the designated goal.

\textbf{OGBench.} To assess the generalization capabilities of our method, we also employ the OGBench suite. Our evaluation includes $50$ state-based tasks, comprising five locomotion and five manipulation environments, each with five distinct dataset compositions. Additionally, we evaluate on five challenging visual manipulation tasks from OGBench. For all OGBench tasks, the primary evaluation metric is the success rate.

\subsection{Methods}

To furnish a rigorous and comprehensive comparative analysis, our experiments are conducted across both offline and online reinforcement learning settings. The selection of baseline algorithms encompasses a diverse spectrum of policy classes and training paradigms, enabling a thorough evaluation of our proposed method.

\textbf{Offline RL Baselines.} In the offline learning context, our empirical comparison is organized around three principal categories of algorithms. For Gaussian policies, we select Implicit Q-Learning (IQL)~\citep{kostrikov2022offline} and Regularized Behavior-Cloning with Active-Critic (ReBRAC)~\citep{tarasov2023rebrac} as strong, standard baselines that employ simple, unimodal policies. To benchmark against diffusion policies, we include IDQL~\citep{hansen-estruch2023idql}, which leverages rejection sampling for policy improvement, and Consistency-AC (CAC)~\citep{ding2024consistency-ac}, which implements policy distillation via backpropagation through a consistency model. Finally, within the domain of flow policies, we benchmark against two distinct strategies: Implicit Flow Q-Learning (IFQL)~\citep{park2025flow}, a flow-based counterpart to IDQL, and Flow Q-Learning (FQL)~\citep{park2025flow}, which represents the state-of-the-art in this class by training an efficient one-step policy via distillation.

\textbf{Offline-to-Online RL Baselines.} For the online fine-tuning phase, our evaluation focuses on algorithms capable of effectively adapting from an offline-pretrained policy. We compare our method against a curated subset of the aforementioned baselines amenable to online interaction, namely IQL, ReBRAC, IFQL, and FQL. This experimental design allows for a direct assessment of both the asymptotic performance and sample efficiency of our approach during online adaptation.

A detailed specification of the hyperparameter configurations for all algorithms and experimental setups is provided in Section \ref{sec:exp_details_methods} to ensure full reproducibility.

\subsection{Results}

\begin{table}[b!]
\vspace{-10pt}
\caption{
\footnotesize
\textbf{Offline RL results.}
DFC achieves the best or near-best performance on most of the 73 diverse, challenging benchmark tasks. 
The performances are averaged over 8 seeds (4 seeds for pixel-based tasks). The best scores are highlighted in \textbf{bold} and the suboptimal scores are labeled as \underline{underlined}. \citet{tarasov2023corl, hansen-estruch2023idql, chen2024srpo} contribute to the cells without the ``$\pm$'' sign.
See Table \ref{table:offline_full} for the full results in Appendix \ref{sec:detailed_results}. 
}
\label{table:offline_selected}
\centering
\scalebox{0.7}
{
\begin{threeparttable}
\begin{tabular}{lccccccc}
\toprule
\texttt{Task Category} & \texttt{IQL} & \texttt{ReBRAC} & \texttt{IDQL} & \texttt{CAC} & \texttt{IFQL} & \texttt{FQL} & \texttt{DFC} \\
\midrule
\texttt{OGBench antmaze-large-singletask ($5$ tasks)} & $53$ {\tiny $\pm 3$} & $\underline{81}$ {\tiny $\pm 5$} & $21$ {\tiny $\pm 5$} & $33$ {\tiny $\pm 4$} & $28$ {\tiny $\pm 5$} & $79$ {\tiny $\pm 3$} & $\mathbf{88}$ {\tiny $\pm 2$} \\
\texttt{OGBench antmaze-giant-singletask ($5$ tasks)} & $4$ {\tiny $\pm 1$} & $\mathbf{26}$ {\tiny $\pm 8$} & $0$ {\tiny $\pm 0$} & $0$ {\tiny $\pm 0$} & $3$ {\tiny $\pm 2$} & $9$ {\tiny $\pm 6$} & $\underline{19} ${\tiny $\pm 14$}\\
\texttt{OGBench humanoidmaze-medium-singletask ($5$ tasks)} & $33$ {\tiny $\pm 2$} & $22$ {\tiny $\pm 8$} & $1$ {\tiny $\pm 0$} & $53$ {\tiny $\pm 8$} & $\underline{60}$ {\tiny $\pm 14$} & $58$ {\tiny $\pm 5$} & $\mathbf{67}$ {\tiny $\pm 14$}\\
\texttt{OGBench humanoidmaze-large-singletask ($5$ tasks)} & $2$ {\tiny $\pm 1$} & $2$ {\tiny $\pm 1$} & $1$ {\tiny $\pm 0$} & $0$ {\tiny $\pm 0$} & $\underline{11}$ {\tiny $\pm 2$} & $4$ {\tiny $\pm 2$} & $\mathbf{19}$ {\tiny $\pm 3$}\\
\texttt{OGBench antsoccer-arena-singletask ($5$ tasks)} & $8$ {\tiny $\pm 2$} & $0$ {\tiny $\pm 0$} & $12$ {\tiny $\pm 4$} & $2$ {\tiny $\pm 4$} & $33$ {\tiny $\pm 6$} & $\underline{60}$ {\tiny $\pm 2$} & $\mathbf{73}$ {\tiny $\pm 4$}\\
\texttt{OGBench cube-single-singletask ($5$ tasks)} & $83$ {\tiny $\pm 3$} & $91$ {\tiny $\pm 2$} & $95$ {\tiny $\pm 2$} & $85$ {\tiny $\pm 9$} & $79$ {\tiny $\pm 2$} & $\underline{96}$ {\tiny $\pm 1$} & $\mathbf{100}$ {\tiny $\pm 0$}\\
\texttt{OGBench cube-double-singletask ($5$ tasks)} & $7$ {\tiny $\pm 1$} & $12$ {\tiny $\pm 1$} & $15$ {\tiny $\pm 6$} & $6$ {\tiny $\pm 2$} & $14$ {\tiny $\pm 3$} & $\underline{29}$ {\tiny $\pm 2$} & $\mathbf{41}$ {\tiny $\pm 4$}\\
\texttt{OGBench scene-singletask ($5$ tasks)} & $28$ {\tiny $\pm 1$} & $41$ {\tiny $\pm 3$} & $46$ {\tiny $\pm 3$} & $40$ {\tiny $\pm 7$} & $30$ {\tiny $\pm 3$} & $\underline{56}$ {\tiny $\pm 2$} & $\mathbf{63}$ {\tiny $\pm 2$}\\
\texttt{OGBench puzzle-3x3-singletask ($5$ tasks)} & $9$ {\tiny $\pm 1$} & $21$ {\tiny $\pm 1$} & $10$ {\tiny $\pm 2$} & $19$ {\tiny $\pm 0$} & $19$ {\tiny $\pm 1$} & $\underline{30}$ {\tiny $\pm 1$} & $\mathbf{40}$ {\tiny $\pm 2$} \\
\texttt{OGBench puzzle-4x4-singletask ($5$ tasks)} & $7$ {\tiny $\pm 1$} & $14$ {\tiny $\pm 1$} & $\underline{29}$ {\tiny $\pm 3$} & $15$ {\tiny $\pm 3$} & $25$ {\tiny $\pm 5$} & $17$ {\tiny $\pm 2$} & $\mathbf{35}$ {\tiny $\pm 4$}\\
\texttt{D4RL antmaze ($6$ tasks)} & $57$ & $78$ & $79$ & $30$ {\tiny $\pm 3$} & $65$ {\tiny $\pm 7$} & $\underline{84}$ {\tiny $\pm 3$} & $\mathbf{92}$ {\tiny $\pm 2$} \\
\texttt{D4RL adroit ($12$ tasks)} & $53$ & $\underline{59}$ & $52$ {\tiny $\pm 1$} & $43$ {\tiny $\pm 2$} & $52$ {\tiny $\pm 1$} & $52$ {\tiny $\pm 1$} & $\mathbf{63}$ {\tiny $\pm 3$}\\
\texttt{Visual manipulation (${5}$ tasks)} & $42$ {\tiny $\pm 4$} & $60$ {\tiny $\pm 2$} & - & - & $50$ {\tiny $\pm 5$} & $\underline{65}$ {\tiny $\pm 2$} & $\mathbf{68}$ {\tiny $\pm 6$}\\
\bottomrule
\end{tabular}
\end{threeparttable}
}
\end{table}

\begin{figure}[h] 
\footnotesize
    \centering 
    \begin{subfigure}{\textwidth} 
        \centering
        \includegraphics[width=\linewidth]{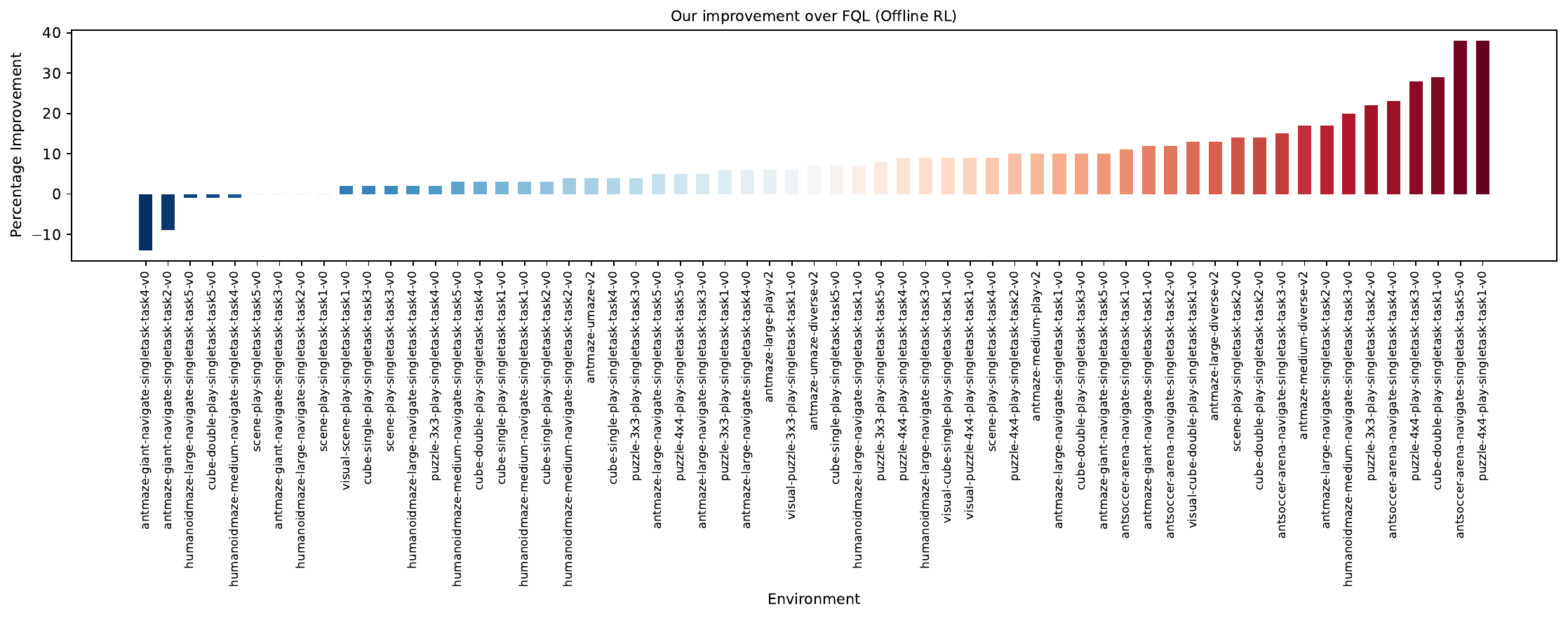}
        \vspace{-1pt}
        \caption{Offline Phase Results Difference}
        \vspace{-1pt}
        \label{fig:offline_results_sub}
    \end{subfigure}
    \begin{subfigure}{\textwidth}
        \centering
        \includegraphics[width=\linewidth]{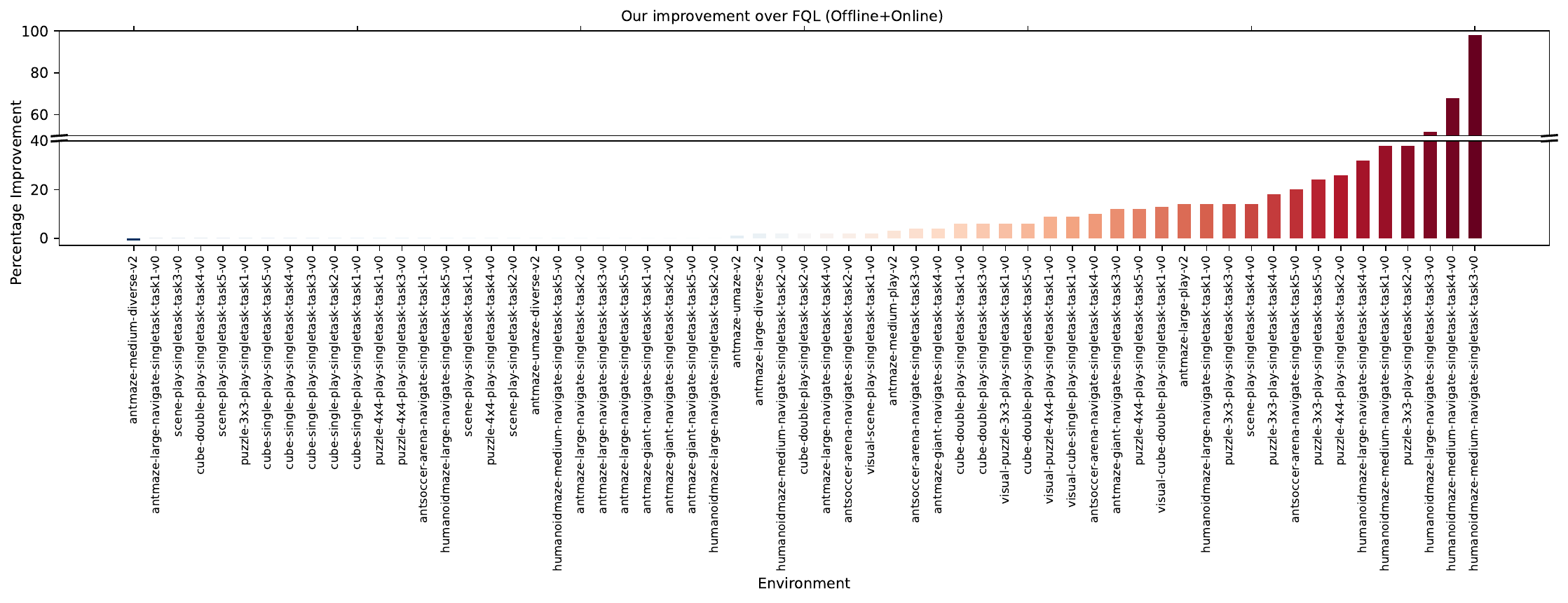}
        \vspace{-1pt}
        \caption{Offline+Online Phase Results Difference}
        \vspace{-1pt}
        \label{fig:online_broken_axis_results_sub}
    \end{subfigure}
    \caption{\textbf{Performance Comparison between DFC and FQL.} Conducted on tasks evaluated by success rates over $8$ seeds.}
    \label{fig:offline_online_comparison}
    \vspace{-10pt}
\end{figure}

Table~\ref{table:offline_selected} presents a comprehensive summary of our aggregated benchmark results across a total of 73 state- and pixel-based offline RL tasks, spanning both robotic locomotion and manipulation. The results indicate that DFC exhibits superior performance over the vast majority of established methods, including those predicated on Gaussian, diffusion, and flow-based policies.

Figure~\ref{fig:offline_online_comparison} further illustrates the performance gains achieved by our method in both offline and offline-to-online settings, demonstrating consistent improvements across most tasks. Our approach substantially outperforms previous methods, especially on some tasks of medium difficulty, while also elevating performance on the majority of other tasks. Particularly noteworthy is DFC's performance on some of the most challenging tasks within the D4RL benchmark, achieving scores of $91\%$ and $95\%$ on \texttt{antmaze-large-play} and \texttt{antmaze-large-diverse}, respectively.

To evaluate DFC's efficacy for fine-tuning, we transition from the offline pre-training phase by incorporating newly collected online transitions into the replay buffer. Following the protocol of FQL, online fine-tuning commences after one million offline gradient steps, continuing to optimize all network components with the same objectives. The evaluation is conducted on the same suite of 68 state-based RL tasks used in our offline experiments.

To underscore the criticality of our proposed architecture, we conduct an ablation study with two simplified variants. The first, a \textbf{Flow-only Critic (FC)}, suffers from significant training instability and high variance across different seeds, rarely surpassing our full model's performance. The second, a \textbf{Distributional-only Critic (DC)}, demonstrates stable training but yields suboptimal performance, offering only marginal improvements over the FQL baseline in limited scenarios. These results, detailed in Table~\ref{table:overall} in Appendix \ref{sec:detailed_results} 
, highlight that while the distributional component ensures stability, it is insufficient on its own, and the combination in our two-stage architecture is essential for achieving state-of-the-art performance.

In contrast to these ablated versions, our full \textbf{DFC} model consistently delivered the best performance. It excelled across a diverse range of challenging domains, including robotic locomotion, manipulation, offline reinforcement learning, and offline-to-online fine-tuning, in both state- and pixel-based settings. Notably, DFC achieves these results without the computational burden of backpropagation through time, highlighting its efficiency and effectiveness.

\section{Conclusion}

In this work, we introduced the Distributional Flow Critic (DFC), a novel critic architecture that synergistically integrates the predictive power of flow matching with the stability of distributional reinforcement learning. Our empirical evaluation demonstrates that this fusion overcomes the training instability inherent in a pure flow-based critic while surpassing the performance limitations of a purely distributional one. The consistent state-of-the-art results across a wide array of challenging benchmarks underscore the efficacy and robustness of our approach. By successfully unifying these two powerful paradigms, DFC establishes a new direction for designing high-performance and stable critics in reinforcement learning. Future work could explore the application of DFC to more complex, partially observable environments or its integration with hierarchical learning frameworks.

\bibliography{main}

\begin{thebibliography}{55}
\providecommand{\natexlab}[1]{#1}
\providecommand{\url}[1]{\texttt{#1}}
\expandafter\ifx\csname urlstyle\endcsname\relax
  \providecommand{\doi}[1]{doi: #1}\else
  \providecommand{\doi}{doi: \begingroup \urlstyle{rm}\Url}\fi

\bibitem[Ada et~al.(2024)Ada, Oztop, and Ugur]{ada2024srdp}
Suzan~Ece Ada, Erhan Oztop, and Emre Ugur.
\newblock Diffusion policies for out-of-distribution generalization in offline reinforcement learning.
\newblock \emph{IEEE Robotics and Automation Letters}, 9\penalty0 (4):\penalty0 3116--3123, 2024.

\bibitem[Adam et~al.(2015)]{kingma2015adam}
Kingma DP Ba~J Adam et~al.
\newblock Adam: A method for stochastic optimization.
\newblock In \emph{International Conference on Learning Representations}, 2015.

\bibitem[Ajay et~al.(2023)Ajay, Du, Gupta, Tenenbaum, Jaakkola, and Agrawal]{ajay2023is}
Anurag Ajay, Yilun Du, Abhi Gupta, Joshua~B. Tenenbaum, Tommi~S. Jaakkola, and Pulkit Agrawal.
\newblock Is conditional generative modeling all you need for decision making?
\newblock In \emph{The Eleventh International Conference on Learning Representations}, 2023.

\bibitem[Alonso et~al.(2024)Alonso, Jelley, Micheli, Kanervisto, Storkey, Pearce, and Fleuret]{alonso2024learning}
Eloi Alonso, Adam Jelley, Vincent Micheli, Anssi Kanervisto, Amos~J Storkey, Tim Pearce, and Fran{\c{c}}ois Fleuret.
\newblock Diffusion for world modeling: Visual details matter in atari.
\newblock \emph{Advances in Neural Information Processing Systems}, 37:\penalty0 58757--58791, 2024.

\bibitem[Barth-Maron et~al.(2018{\natexlab{a}})Barth-Maron, Hoffman, Budden, Dabney, Horgan, TB, Muldal, Heess, and Lillicrap]{barth-maron2018distributed}
Gabriel Barth-Maron, Matthew~W. Hoffman, David Budden, Will Dabney, Dan Horgan, Dhruva TB, Alistair Muldal, Nicolas Heess, and Timothy Lillicrap.
\newblock Distributional policy gradients.
\newblock In \emph{International Conference on Learning Representations}, 2018{\natexlab{a}}.

\bibitem[Barth-Maron et~al.(2018{\natexlab{b}})Barth-Maron, Hoffman, Budden, Dabney, Horgan, TB, Muldal, Heess, and Lillicrap]{barth2018distributed}
Gabriel Barth-Maron, Matthew~W Hoffman, David Budden, Will Dabney, Dan Horgan, Dhruva TB, Alistair Muldal, Nicolas Heess, and Timothy Lillicrap.
\newblock Distributed distributional deterministic policy gradients.
\newblock In \emph{International Conference on Learning Representations}, 2018{\natexlab{b}}.

\bibitem[Bellemare et~al.(2017)Bellemare, Dabney, and Munos]{bellemare2017distributional}
Marc~G. Bellemare, Will Dabney, and R{\'e}mi Munos.
\newblock A distributional perspective on reinforcement learning.
\newblock In \emph{Proceedings of the 34th International Conference on Machine Learning (ICML)}, pp.\  449--458, 2017.

\bibitem[Chen et~al.(2024{\natexlab{a}})Chen, Deng, Kawaguchi, Gulcehre, and Ahn]{chen2024hierarchical}
Chang Chen, Fei Deng, Kenji Kawaguchi, Caglar Gulcehre, and Sungjin Ahn.
\newblock Simple hierarchical planning with diffusion.
\newblock In \emph{The Twelfth International Conference on Learning Representations}, 2024{\natexlab{a}}.

\bibitem[Chen et~al.(2023)Chen, Lu, Ying, Su, and Zhu]{chen2023sfbc}
Huayu Chen, Cheng Lu, Chengyang Ying, Hang Su, and Jun Zhu.
\newblock Offline reinforcement learning via high-fidelity generative behavior modeling.
\newblock In \emph{The Eleventh International Conference on Learning Representations}, 2023.

\bibitem[Chen et~al.(2024{\natexlab{b}})Chen, Lu, Wang, Su, and Zhu]{chen2024srpo}
Huayu Chen, Cheng Lu, Zhengyi Wang, Hang Su, and Jun Zhu.
\newblock Score regularized policy optimization through diffusion behavior.
\newblock In \emph{The Twelfth International Conference on Learning Representations}, 2024{\natexlab{b}}.

\bibitem[Chen et~al.(2024{\natexlab{c}})Chen, Wang, and Zhou]{chen2024dtql}
Tianyu Chen, Zhendong Wang, and Mingyuan Zhou.
\newblock Diffusion policies creating a trust region for offline reinforcement learning.
\newblock \emph{Advances in Neural Information Processing Systems}, 37:\penalty0 50098--50125, 2024{\natexlab{c}}.

\bibitem[Dabney et~al.(2018{\natexlab{a}})Dabney, Ostrovski, Silver, and Munos]{dabney2018implicit}
Will Dabney, Georg Ostrovski, David Silver, and R{\'e}mi Munos.
\newblock Implicit quantile networks for distributional reinforcement learning.
\newblock In \emph{International conference on machine learning}, pp.\  1096--1105. PMLR, 2018{\natexlab{a}}.

\bibitem[Dabney et~al.(2018{\natexlab{b}})Dabney, Rowland, Bellemare, and Munos]{dabney2018distributional}
Will Dabney, Mark Rowland, Marc~G. Bellemare, and R{\'e}mi Munos.
\newblock Distributional reinforcement learning with quantile regression.
\newblock In \emph{AAAI Conference on Artificial Intelligence}, volume~32, 2018{\natexlab{b}}.

\bibitem[Ding \& Jin(2024{\natexlab{a}})Ding and Jin]{ding2024consistency-ac}
Zihan Ding and Chi Jin.
\newblock Consistency models as a rich and efficient policy class for reinforcement learning.
\newblock In \emph{The Twelfth International Conference on Learning Representations}, 2024{\natexlab{a}}.

\bibitem[Ding \& Jin(2024{\natexlab{b}})Ding and Jin]{ding2024consistencyac}
Zihan Ding and Chi Jin.
\newblock Consistency models as a rich and efficient policy class for reinforcement learning.
\newblock In \emph{The Twelfth International Conference on Learning Representations}, 2024{\natexlab{b}}.

\bibitem[Ding et~al.(2024)Ding, Zhang, Tian, and Zheng]{ding2024diffusion}
Zihan Ding, Amy Zhang, Yuandong Tian, and Qinqing Zheng.
\newblock Diffusion world model.
\newblock \emph{CoRR}, 2024.

\bibitem[Espeholt et~al.(2018)Espeholt, Soyer, Munos, Simonyan, Mnih, Ward, Doron, Firoiu, Harley, Dunning, et~al.]{espeholt2018impala}
Lasse Espeholt, Hubert Soyer, Remi Munos, Karen Simonyan, Vlad Mnih, Tom Ward, Yotam Doron, Vlad Firoiu, Tim Harley, Iain Dunning, et~al.
\newblock Impala: Scalable distributed deep-rl with importance weighted actor-learner architectures.
\newblock In \emph{International conference on machine learning}, pp.\  1407--1416. PMLR, 2018.

\bibitem[Esser et~al.(2024)Esser, Kulal, Blattmann, Entezari, M{\"u}ller, Saini, Levi, Lorenz, Sauer, Boesel, et~al.]{esser2024scaling}
Patrick Esser, Sumith Kulal, Andreas Blattmann, Rahim Entezari, Jonas M{\"u}ller, Harry Saini, Yam Levi, Dominik Lorenz, Axel Sauer, Frederic Boesel, et~al.
\newblock Scaling rectified flow transformers for high-resolution image synthesis.
\newblock In \emph{Forty-first international conference on machine learning}, 2024.

\bibitem[Fu et~al.(2020)Fu, Kumar, Nachum, Tucker, and Levine]{fu2020d4rl}
Justin Fu, Aviral Kumar, Ofir Nachum, George Tucker, and Sergey Levine.
\newblock D4rl: Datasets for deep data-driven reinforcement learning.
\newblock \emph{arXiv preprint arXiv:2004.07219}, 2020.

\bibitem[Fujimoto \& Gu(2021)Fujimoto and Gu]{fujimoto2021minimalist}
Scott Fujimoto and Shixiang~Shane Gu.
\newblock A minimalist approach to offline reinforcement learning.
\newblock \emph{Advances in neural information processing systems}, 34:\penalty0 20132--20145, 2021.

\bibitem[Hansen-Estruch et~al.(2023)Hansen-Estruch, Kostrikov, Janner, Kuba, and Levine]{hansen-estruch2023idql}
Philippe Hansen-Estruch, Ilya Kostrikov, Michael Janner, Jakub~Grudzien Kuba, and Sergey Levine.
\newblock Idql: Implicit q-learning as an actor-critic method with diffusion policies.
\newblock \emph{arXiv preprint arXiv:2304.10573}, 2023.

\bibitem[He et~al.(2023)He, Shen, Zhang, Tan, and Wang]{he2023diffcps}
Longxiang He, Li~Shen, Linrui Zhang, Junbo Tan, and Xueqian Wang.
\newblock Diffcps: Diffusion model based constrained policy search for offline reinforcement learning.
\newblock \emph{arXiv preprint arXiv:2310.05333}, 2023.

\bibitem[He et~al.(2024)He, Shen, Tan, and Wang]{he2024aligniql}
Longxiang He, Li~Shen, Junbo Tan, and Xueqian Wang.
\newblock Aligniql: Policy alignment in implicit q-learning through constrained optimization.
\newblock \emph{arXiv preprint arXiv:2405.18187}, 2024.

\bibitem[Hendrycks \& Gimpel(2016)Hendrycks and Gimpel]{hendrycks2016gelu}
Dan Hendrycks and Kevin Gimpel.
\newblock Gaussian error linear units (gelus).
\newblock \emph{arXiv preprint arXiv:1606.08415}, 2016.

\bibitem[Ho et~al.(2020)Ho, Jain, and Abbeel]{ho2020denoising}
Jonathan Ho, Ajay Jain, and Pieter Abbeel.
\newblock Denoising diffusion probabilistic models.
\newblock \emph{Advances in neural information processing systems}, 33:\penalty0 6840--6851, 2020.

\bibitem[Jackson et~al.(2024)Jackson, Matthews, Lu, Ellis, Whiteson, and Foerster]{jackson2024policy}
Matthew~Thomas Jackson, Michael Matthews, Cong Lu, Benjamin Ellis, Shimon Whiteson, and Jakob~Nicolaus Foerster.
\newblock Policy-guided diffusion.
\newblock In \emph{Reinforcement Learning Conference}, 2024.

\bibitem[Janner et~al.(2022)Janner, Du, Tenenbaum, and Levine]{janner2022planning}
Michael Janner, Yilun Du, Joshua Tenenbaum, and Sergey Levine.
\newblock Planning with diffusion for flexible behavior synthesis.
\newblock In \emph{International Conference on Machine Learning}, pp.\  9902--9915. PMLR, 2022.

\bibitem[Kang et~al.(2023)Kang, Ma, Du, Pang, and Yan]{kang2023edp}
Bingyi Kang, Xiao Ma, Chao Du, Tianyu Pang, and Shuicheng Yan.
\newblock Efficient diffusion policies for offline reinforcement learning.
\newblock \emph{Advances in Neural Information Processing Systems}, 36:\penalty0 67195--67212, 2023.

\bibitem[Koenker \& Bassett(1978)Koenker and Bassett]{koenker1978regression}
Roger Koenker and Jr. Bassett, Gilbert.
\newblock Regression quantiles.
\newblock \emph{Econometrica}, 46\penalty0 (1):\penalty0 33--50, 1978.

\bibitem[Kostrikov et~al.(2022)Kostrikov, Nair, and Levine]{kostrikov2022offline}
Ilya Kostrikov, Ashvin Nair, and Sergey Levine.
\newblock Offline reinforcement learning with implicit q-learning.
\newblock In \emph{International Conference on Learning Representations}, 2022.

\bibitem[Li et~al.(2024)Li, Krohn, Chen, Ajay, Agrawal, and Chalvatzaki]{li2024ddiffpg}
Steven Li, Rickmer Krohn, Tao Chen, Anurag Ajay, Pulkit Agrawal, and Georgia Chalvatzaki.
\newblock Learning multimodal behaviors from scratch with diffusion policy gradient.
\newblock \emph{Advances in Neural Information Processing Systems}, 37:\penalty0 38456--38479, 2024.

\bibitem[Liang et~al.(2023)Liang, Mu, Ding, Ni, Tomizuka, and Luo]{liang2023adaptdiffuser}
Zhixuan Liang, Yao Mu, Mingyu Ding, Fei Ni, Masayoshi Tomizuka, and Ping Luo.
\newblock Adaptdiffuser: diffusion models as adaptive self-evolving planners.
\newblock In \emph{Proceedings of the 40th International Conference on Machine Learning}, pp.\  20725--20745, 2023.

\bibitem[Lipman et~al.(2023)Lipman, Chen, Ben-Hamu, Nickel, and Le]{lipman2023flow}
Yaron Lipman, Ricky~TQ Chen, Heli Ben-Hamu, Maximilian Nickel, and Matt Le.
\newblock Flow matching for generative modeling.
\newblock In \emph{International Conference on Learning Representations}, 2023.

\bibitem[Lu et~al.(2023{\natexlab{a}})Lu, Chen, Chen, Su, Li, and Zhu]{lu2023qgpo}
Cheng Lu, Huayu Chen, Jianfei Chen, Hang Su, Chongxuan Li, and Jun Zhu.
\newblock Contrastive energy prediction for exact energy-guided diffusion sampling in offline reinforcement learning.
\newblock In \emph{International Conference on Machine Learning}, pp.\  22825--22855. PMLR, 2023{\natexlab{a}}.

\bibitem[Lu et~al.(2023{\natexlab{b}})Lu, Ball, Teh, and Parker-Holder]{lu2023synther}
Cong Lu, Philip Ball, Yee~Whye Teh, and Jack Parker-Holder.
\newblock Synthetic experience replay.
\newblock \emph{Advances in Neural Information Processing Systems}, 36:\penalty0 46323--46344, 2023{\natexlab{b}}.

\bibitem[Mazoure et~al.(2020)Mazoure, Doan, Durand, Pineau, and Hjelm]{mazoure2020leveraging}
Bogdan Mazoure, Thang Doan, Audrey Durand, Joelle Pineau, and R~Devon Hjelm.
\newblock Leveraging exploration in off-policy algorithms via normalizing flows.
\newblock In \emph{Conference on Robot Learning}, pp.\  430--444. PMLR, 2020.

\bibitem[Morimura et~al.(2010)Morimura, Sugiyama, Kashima, Hachiya, and Tanaka]{morimura2010parametric}
Tetsuro Morimura, Masashi Sugiyama, Hisashi Kashima, Hirotaka Hachiya, and Toshiyuki Tanaka.
\newblock Parametric return density estimation for reinforcement learning.
\newblock In \emph{Conference on Uncertainty in Artificial Intelligence}, 2010.

\bibitem[Park et~al.(2025{\natexlab{a}})Park, Frans, Eysenbach, and Levine]{park2025ogbench}
Seohong Park, Kevin Frans, Benjamin Eysenbach, and Sergey Levine.
\newblock {OGB}ench: Benchmarking offline goal-conditioned {RL}.
\newblock In \emph{The Thirteenth International Conference on Learning Representations}, 2025{\natexlab{a}}.

\bibitem[Park et~al.(2025{\natexlab{b}})Park, Frans, Eysenbach, and Levine]{parkogbench}
Seohong Park, Kevin Frans, Benjamin Eysenbach, and Sergey Levine.
\newblock Ogbench: Benchmarking offline goal-conditioned rl.
\newblock In \emph{The Thirteenth International Conference on Learning Representations}, 2025{\natexlab{b}}.

\bibitem[Park et~al.(2025{\natexlab{c}})Park, Li, and Levine]{park2025flow}
Seohong Park, Qiyang Li, and Sergey Levine.
\newblock Flow q-learning.
\newblock In \emph{Forty-second International Conference on Machine Learning}, 2025{\natexlab{c}}.

\bibitem[Psenka et~al.(2024)Psenka, Escontrela, Abbeel, and Ma]{psenka2024qsm}
Michael Psenka, Alejandro Escontrela, Pieter Abbeel, and Yi~Ma.
\newblock Learning a diffusion model policy from rewards via q-score matching.
\newblock In \emph{Proceedings of the 41st International Conference on Machine Learning}, pp.\  41163--41182, 2024.

\bibitem[Ren et~al.(2024)Ren, Lidard, Ankile, Simeonov, Agrawal, Majumdar, Burchfiel, Dai, and Simchowitz]{ren2024diffusion}
Allen~Z. Ren, Justin Lidard, Lars~Lien Ankile, Anthony Simeonov, Pulkit Agrawal, Anirudha Majumdar, Benjamin Burchfiel, Hongkai Dai, and Max Simchowitz.
\newblock Diffusion policy policy optimization.
\newblock In \emph{CoRL 2024 Workshop on Mastering Robot Manipulation in a World of Abundant Data}, 2024.

\bibitem[Singh et~al.(2022)Singh, Lee, and Chen]{singh2022sample-based}
Rahul Singh, Keuntaek Lee, and Yongxin Chen.
\newblock Sample-based distributional policy gradient.
\newblock In \emph{Learning for Dynamics and Control Conference}, pp.\  676--688. PMLR, 2022.

\bibitem[Sohl-Dickstein et~al.(2015)Sohl-Dickstein, Weiss, Maheswaranathan, and Ganguli]{sohl2015deep}
Jascha Sohl-Dickstein, Eric Weiss, Niru Maheswaranathan, and Surya Ganguli.
\newblock Deep unsupervised learning using nonequilibrium thermodynamics.
\newblock In \emph{International conference on machine learning}, pp.\  2256--2265. pmlr, 2015.

\bibitem[Suh et~al.(2023)Suh, Chou, Dai, Yang, Gupta, and Tedrake]{suh2023skill}
HJ~Terry Suh, Glen Chou, Hongkai Dai, Lujie Yang, Abhishek Gupta, and Russ Tedrake.
\newblock Fighting uncertainty with gradients: Offline reinforcement learning via diffusion score matching.
\newblock In \emph{Conference on Robot Learning}, pp.\  2878--2904. PMLR, 2023.

\bibitem[Sutton \& Barto(2018)Sutton and Barto]{sutton2018mdp}
Richard~S. Sutton and Andrew~G. Barto.
\newblock \emph{Finite Markov Decision Processes}, pp.\  67--69.
\newblock MIT Press, 2nd edition, 2018.

\bibitem[Tarasov et~al.(2023{\natexlab{a}})Tarasov, Kurenkov, Nikulin, and Kolesnikov]{tarasov2023rebrac}
Denis Tarasov, Vladislav Kurenkov, Alexander Nikulin, and Sergey Kolesnikov.
\newblock Revisiting the minimalist approach to offline reinforcement learning.
\newblock \emph{Advances in Neural Information Processing Systems}, 36:\penalty0 11592--11620, 2023{\natexlab{a}}.

\bibitem[Tarasov et~al.(2023{\natexlab{b}})Tarasov, Nikulin, Akimov, Kurenkov, and Kolesnikov]{tarasov2023corl}
Denis Tarasov, Alexander Nikulin, Dmitry Akimov, Vladislav Kurenkov, and Sergey Kolesnikov.
\newblock Corl: Research-oriented deep offline reinforcement learning library.
\newblock \emph{Advances in Neural Information Processing Systems}, 36:\penalty0 30997--31020, 2023{\natexlab{b}}.

\bibitem[Venkatraman et~al.(2024)Venkatraman, Khaitan, Akella, Dolan, Schneider, and Berseth]{venkatraman2024learning}
Siddarth Venkatraman, Shivesh Khaitan, Ravi~Tej Akella, John Dolan, Jeff Schneider, and Glen Berseth.
\newblock Reasoning with latent diffusion in offline reinforcement learning.
\newblock In \emph{The Twelfth International Conference on Learning Representations}, 2024.

\bibitem[Villani(2003)]{villani2003topics}
C{\'e}dric Villani.
\newblock \emph{Topics in Optimal Transportation}, volume~58.
\newblock American Mathematical Society, 2003.

\bibitem[Wang et~al.(2023)Wang, Hunt, and Zhou]{wang2023dql}
Zhendong Wang, Jonathan~J Hunt, and Mingyuan Zhou.
\newblock Diffusion policies as an expressive policy class for offline reinforcement learning.
\newblock In \emph{The Eleventh International Conference on Learning Representations}, 2023.

\bibitem[Yang et~al.(2023)Yang, Huang, Lei, Zhong, Yang, Fang, Wen, Zhou, and Lin]{yang2023dipo}
Long Yang, Zhixiong Huang, Fenghao Lei, Yucun Zhong, Yiming Yang, Cong Fang, Shiting Wen, Binbin Zhou, and Zhouchen Lin.
\newblock Policy representation via diffusion probability model for reinforcement learning.
\newblock \emph{arXiv preprint arXiv:2305.13122}, 2023.

\bibitem[Zhang et~al.(2024)Zhang, Luo, Sj{\"o}lund, Sch{\"o}n, and Mattsson]{zhang2024entropydql}
Ruoqi Zhang, Ziwei Luo, Jens Sj{\"o}lund, Thomas Sch{\"o}n, and Per Mattsson.
\newblock Entropy-regularized diffusion policy with q-ensembles for offline reinforcement learning.
\newblock \emph{Advances in Neural Information Processing Systems}, 37:\penalty0 98871--98897, 2024.

\bibitem[Zhang et~al.(2025)Zhang, Zhang, and Gu]{zhang2025energy}
Shiyuan Zhang, Weitong Zhang, and Quanquan Gu.
\newblock Energy-weighted flow matching for offline reinforcement learning.
\newblock In \emph{International Conference on Learning Representations (ICLR)}, 2025.

\bibitem[Zheng et~al.(2023)Zheng, Le, Shaul, Lipman, Grover, and Chen]{zheng2023guided}
Qinqing Zheng, Matt Le, Neta Shaul, Yaron Lipman, Aditya Grover, and Ricky T.~Q. Chen.
\newblock Guided flows for generative modeling and decision making.
\newblock \emph{CoRR}, abs/2311.13443, 2023.

\end{thebibliography}
\bibliographystyle{bst}
\clearpage
\appendix

\section{Implementation Details}\label{sec:implementation_details}
\subsection{Hyperparameters and Architecture}\label{sec:exp_details_methods}
Across all experiments, we set the learning rates for the critic network $Z_\phi$ and $Z_{\tilde{\phi}}$ to be $3 \times 10^{-4}$ and $1 \times 10^{-4}$. We use a sample size of $M = 51$ for the distributional estimation (following the classical setting migrated from \citet{bellemare2017distributional}), an exploration constant of $\delta = 0.3$, a Huber loss parameter of $\zeta = 1$, and 10 integration steps for the flow models. For all neural network components, which are realized as $[512, 512, 512, 512]$-sized multi-layer perceptions (MLPs), we follow the architectural configuration of FQL. To ensure a fair and direct comparison, we also adopt the identical hyperparameters for the actor, including the behavioral cloning coefficient $\alpha_{\text{BC}}$, as specified in FQL. We provide the complete list of hyperparameters in Table \ref{table:hyp}.

Due to computational resource constraints, our hyperparameter selection largely adheres to the configurations established in FQL. A comprehensive list of general hyperparameters is provided in Table \ref{table:hyp}, while task-specific settings are detailed in Table \ref{table:method_hyp}. For key baseline methods, we made the following specific adjustments:

\begin{itemize}[leftmargin=*]
    \item \textbf{For IQL and IFQL}, we set exploring expectile value to be $0.9$, while the AWR inverse temperature, $\alpha$, is tuned on a per-environment basis (see Table \ref{table:method_hyp}). 
    
    \item \textbf{For ReBRAC}, the actor BC coefficient, $\alpha_1$, and the critic BC coefficient, $\alpha_2$, are set on a per-environment basis, as specified in Table \ref{table:method_hyp}.
    
    \item \textbf{For IDQL}, the expectile value, $\tau$, is configured to 0.7 for OGBench locomotion and Adroit tasks, and to 0.9 for OGBench manipulation and AntMaze tasks. The number of action samples, $N$, is also tuned individually per task (see Table \ref{table:method_hyp}). Consistent with its original protocol, the agent is trained for 3 million steps (1.5 million for the value function).
    
    \item \textbf{For CAC}, the Q-loss coefficient, $\eta$, is determined based on the environment-specific values listed in Table \ref{table:method_hyp}.
    
\end{itemize}

\begin{table}[h!]
\caption{
\footnotesize
\textbf{Hyperparameters for DFC.}
$(*)$ Denotes hyperparameters for which the values are kept identical to the FQL configuration to ensure a fair comparison.
}
\label{table:hyp}
\begin{center}
\scalebox{0.8}
{
\begin{tabular}{ll}
    \toprule
    \textbf{Hyperparameter} & \textbf{Value} \\
    \midrule
    Distributional critic learning rate & $0.0001$\\
    Main critic learning rate & $0.0003$\\
    Distributional sample size & $50$\\
    Exploration constant $\delta$ & $0.3$\\
    Huber loss parameter $\zeta$ & $1$\\
    Actor learning rate $(*)$ & $0.0003$ \\
    Optimizer $(*)$& Adam~\citep{kingma2015adam} \\
    Gradient steps $(*)$& $1000000$ (default), $500000$ (Offline D4RL, pixel-based OGBench) \\
    Minibatch size $(*)$& $256$ \\
    MLP dimensions $(*)$& $[512, 512, 512, 512]$ \\
    Nonlinearity $(*)$& GELU~\citep{hendrycks2016gelu} \\
    Target network smoothing coefficient $(*)$& $0.005$ \\
    Discount factor $\gamma$ $(*)$& $0.99$ (default), $0.995$ (\texttt{antmaze-giant}, \texttt{humanoidmaze}, \texttt{antsoccer}) \\
    Image augmentation probability $(*)$& $0.5$ \\
    Flow steps & $10$ \\
    Flow time sampling distribution & $\mathcal{U}([0, 1])$ \\
    BC coefficient $\alpha$ $(*)$& See Table \ref{table:method_hyp}. \\
    \bottomrule
\end{tabular}
}
\end{center}
\end{table} 


For the state-based OGBench tasks, DFC is trained for one million gradient steps, whereas for the D4RL and pixel-based OGBench tasks, training is conducted for 500K steps, in alignment with the FQL protocol. The agent's performance is evaluated every 100K steps over 50 episodes. For the offline-to-online experiments, we report the performance metrics at both one million and two million total steps. For reporting final scores on OGBench, we follow the protocol of FQL and average the success rates over the last three evaluation epochs (i.e., 800K, 900K, and 1M steps for state-based tasks; 300K, 400K, and 500K for pixel-based tasks). For D4RL, consistent with \citet{tarasov2023corl}, we report the performance at the final training epoch.

In accordance with the official implementation of OGBench, for all pixel-based experiments, we use a compact version of the IMPALA encoder~\citep{espeholt2018impala} with random random-shift augmentation.

\begin{table}[h!]
\caption{
\footnotesize
\textbf{Task-specific hyperparameters.} 
For a detailed description of each hyperparameter, we refer the reader to Appendix \ref{sec:exp_details_methods}. Some task-specific hyperparameter values are adopted from FQL. To ensure experimental consistency, a single hyperparameter configuration is applied uniformly to all five tasks derived from each OGBench environment. An em-dash ``-" indicates that a corresponding result is not available.
}
\label{table:method_hyp}
\centering
\vspace{-5pt}
\scalebox{0.66}
{
\newlength{\mylen}
\setlength{\mylen}{19pt}
\begin{tabular}{l@{\hskip \mylen}c@{\hskip \mylen}c@{\hskip \mylen}c@{\hskip \mylen}c@{\hskip \mylen}c@{\hskip \mylen}c}
\toprule
& \texttt{IQL} & \texttt{ReBRAC} & \texttt{IDQL} & \texttt{CAC} & \texttt{IFQL} & \texttt{DFC (FQL)} \\
\texttt{Task} & $\alpha$ & $(\alpha_1, \alpha_2)$ & $N$ & $\eta$ & $N$ & {$\alpha$} \\
\midrule
\texttt{antmaze-large-navigate-singletask-v0} & $10$ & $(0.003, 0.01)$ & $32$ & $1$ & $32$ & $10$ \\
\midrule
\texttt{antmaze-giant-navigate-singletask-v0} & $10$ & $(0.003, 0.01)$ & $32$ & $1$ & $32$ & $10$ \\
\midrule
\texttt{humanoidmaze-medium-navigate-singletask-v0} & $10$ & $(0.01, 0.01)$ & $32$ & $0.03$ & $32$ & $30$ \\
\midrule
\texttt{humanoidmaze-large-navigate-singletask-v0} & $10$ & $(0.01, 0.01)$ & $32$ & $1$ & $32$ & $30$ \\
\midrule
\texttt{antsoccer-arena-navigate-singletask-v0} & $1$ & $(0.01, 0.01)$ & $32$ & $1$ & $64$ & $10$ \\
\midrule
\texttt{cube-single-play-singletask-v0} & $1$ & $(1, 0)$ & $32$ & $0.003$ & $32$ & $300$ \\
\midrule
\texttt{cube-double-play-singletask-v0} & $0.3$ & $(0.1, 0)$ & $32$ & $0.3$ & $32$ & $300$ \\
\midrule
\texttt{scene-play-singletask-v0} & $10$ & $(0.1, 0.01)$ & $32$ & $0.3$ & $32$ & $300$ \\
\midrule
\texttt{puzzle-3x3-play-singletask-v0} & $10$ & $(0.3, 0.01)$ & $32$ & $0.01$ & $32$ & $1000$ \\
\midrule
\texttt{puzzle-4x4-play-singletask-v0} & $3$ & $(0.3, 0.01)$ & $32$ & $0.01$ & $32$ & $1000$ \\
\midrule
\texttt{antmaze-umaze-v2} & 10 & $(0.003, 0.002)$ & $32$ & $0.01$ & $32$ & $10$ \\
\texttt{antmaze-umaze-diverse-v2} & 10 &$(0.003, 0.001)$ & $32$ & $0.01$ & $32$ & $10$ \\
\texttt{antmaze-medium-play-v2} & 10 & $(0.001, 0.0005)$ & $32$ & $0.01$ & $32$ & $10$ \\
\texttt{antmaze-medium-diverse-v2} & 10 & $(0.001, 0.0)$ & $32$ & $0.01$ & $32$ & $10$ \\
\texttt{antmaze-large-play-v2} & 10 & $(0.002, 0.001)$ & $32$ & $4.5$ & $32$ & $3$ \\
\texttt{antmaze-large-diverse-v2} & 10 & $(0.002, 0.002)$ & $32$ & $3.5$ & $32$ & $3$ \\
\midrule
\texttt{pen-human-v1} & 3 & $(0.1, 0.5)$ & $32$ & $0.003$ & $32$ & $10000$ \\
\texttt{pen-cloned-v1} & 3 & $(0.05, 0.5)$ & $32$ & $0.003$ & $32$ & $10000$ \\
\texttt{pen-expert-v1} & 3 & $(0.01, 0.01)$ & $32$ & $0.03$ & $32$ & $3000$ \\
\texttt{door-human-v1} & 3 & $(0.1, 0.1)$ & $32$ & $0.03$ & $32$ & $30000$ \\
\texttt{door-cloned-v1} & 3 & $(0.01, 0.1)$ & $32$ & $0.03$ & $128$ & $30000$ \\
\texttt{door-expert-v1} & 3 & $(0.05, 0.01)$ & $32$ & $0.03$ & $32$ & $30000$ \\
\texttt{hammer-human-v1} & 3 & $(0.01, 0.5)$ & $128$ & $0.03$ & $32$ & $30000$ \\
\texttt{hammer-cloned-v1} & 3 & $(0.1, 0.5)$ & $32$ & $0.003$ & $32$ & $10000$ \\
\texttt{hammer-expert-v1} & 3 & $(0.01, 0.01)$ & $32$ & $0.03$ & $32$ & $30000$ \\
\texttt{relocate-human-v1} & 3 & $(0.1, 0.01)$ & $32$ & $0.01$ & $128$ & $10000$ \\
\texttt{relocate-cloned-v1} & 3 & $(0.1, 0.01)$ & $64$ & $0.01$ & $32$ & $30000$ \\
\texttt{relocate-expert-v1} & 3 & $(0.05, 0.01)$ & $32$ & $0.003$ & $32$ & $30000$ \\
\midrule
\texttt{visual-cube-single-play-singletask-task1-v0} & $1$ & $(1, 0)$ & - & - & $32$ & $300$ \\
\texttt{visual-cube-double-play-singletask-task1-v0} & $0.3$ & $(0.1, 0)$ & - & - & $32$ & $100$ \\
\texttt{visual-scene-play-singletask-task1-v0} & $10$ & $(0.1, 0.01)$ & - & - & $32$ & $100$ \\
\texttt{visual-puzzle-3x3-play-singletask-task1-v0} & $10$ & $(0.3, 0.01)$ & - & - & $32$ & $300$ \\
\texttt{visual-puzzle-4x4-play-singletask-task1-v0} & $3$ & $(0.3, 0.01)$ & - & - & $32$ & $300$ \\
\bottomrule
\end{tabular}
}
\vspace{-10pt}
\end{table}

\subsection{Datasets and Environments}

Our empirical evaluation is conducted on two prominent offline reinforcement learning benchmarks: OGBench and D4RL. These benchmarks provide a diverse set of environments and tasks designed to test the limits of offline RL algorithms, particularly their ability to stitch together suboptimal trajectories into effective policies. Detailed results are provided in Table \ref{table:offline_full}.

\textbf{OGBench}~\citep{park2025ogbench} features a semi-sparse reward structure where the reward is defined as the negative count of remaining subtasks required to achieve a fixed goal. This structure presents distinct challenges: locomotion tasks typically involve a single subtask (reaching a goal) with binary rewards ($0$ or $-1$), while manipulation tasks can involve up to $16$ sequential subtasks. The datasets are collected from policies executing random tasks, resulting in highly suboptimal data that necessitates strong stitching capabilities from the learning algorithm.

Our experiments leverage the following OGBench environments, categorized by input type and task domain:

\begin{itemize}[leftmargin=*]
    \item \textbf{State-Based Tasks (50 tasks from 10 environments):} These tasks require control based on proprioceptive states.
    \begin{itemize}
        \item \textit{Locomotion}: The agent, either a quadruped (\texttt{ant}) or a humanoid, must navigate through various maze layouts or dribble a ball to a target location.
        \begin{itemize}[label=\textbullet]
            \item \texttt{antmaze-large-navigate-v0}
            \item \texttt{antmaze-giant-navigate-v0}
            \item \texttt{humanoidmaze-medium-navigate-v0}
            \item \texttt{humanoidmaze-large-navigate-v0}
            \item \texttt{antsoccer-arena-navigate-v0}
        \end{itemize}
        \item \textit{Manipulation}: A robotic arm must manipulate diverse objects. These tasks range from simple object interaction (\texttt{cube}) to long-horizon, multi-object control (\texttt{scene}) and challenges requiring combinatorial generalization (\texttt{puzzle}).
        \begin{itemize}[label=\textbullet]
            \item \texttt{cube-single-play-v0}
            \item \texttt{cube-double-play-v0}
            \item \texttt{scene-play-v0}
            \item \texttt{puzzle-3x3-play-v0}
            \item \texttt{puzzle-4x4-play-v0}
        \end{itemize}
    \end{itemize}
    \item \textbf{Pixel-Based Tasks (5 tasks from 5 environments):} These are the visual counterparts to the manipulation tasks, requiring control solely from $64 \times 64 \times 3$ image observations.
    \begin{itemize}[label=\textbullet]
        \item \texttt{visual-cube-single-play-v0}
        \item \texttt{visual-cube-double-play-v0}
        \item \texttt{visual-scene-play-v0}
        \item \texttt{visual-puzzle-3x3-play-v0}
        \item \texttt{visual-puzzle-4x4-play-v0}
    \end{itemize}
\end{itemize}

\textbf{D4RL}~\citep{fu2020d4rl} is also evaluated on $18$ challenging tasks to facilitate direct comparison with a broad range of prior work. These tasks are selected to cover complex locomotion and dexterous manipulation challenges.

\begin{itemize}[leftmargin=*]
    \item \textbf{AntMaze Tasks (6 tasks):} These locomotion tasks share a similar high-level objective with their OGBench counterparts but feature different maze layouts and dataset characteristics.
    \begin{itemize}[label=\textbullet]
        \item \texttt{antmaze-umaze-v2}
        \item \texttt{antmaze-umaze-diverse-v2}
        \item \texttt{antmaze-medium-play-v2}
        \item \texttt{antmaze-medium-diverse-v2}
        \item \texttt{antmaze-large-play-v2}
        \item \texttt{antmaze-large-diverse-v2}
    \end{itemize}
    \item \textbf{Adroit Tasks (12 tasks):} These tasks demand dexterous manipulation using a high-dimensional ($24$-DoF) robotic hand, with objectives such as spinning a pen, opening a door, hammering a nail, and relocating an object.
    \begin{itemize}[label=\textbullet]
        \item \texttt{pen-human-v1}, \texttt{-cloned-v1}, \texttt{-expert-v1}
        \item \texttt{door-human-v1}, \texttt{-cloned-v1}, \texttt{-expert-v1}
        \item \texttt{hammer-human-v1}, \texttt{-cloned-v1}, \texttt{-expert-v1}
        \item \texttt{relocate-human-v1}, \texttt{-cloned-v1}, \texttt{-expert-v1}
    \end{itemize}
\end{itemize}

Evaluation metrics adhere to the standard protocols for each benchmark: success rates for OGBench and D4RL AntMaze tasks, and normalized returns for Adroit tasks.


\section{Detailed Results}\label{sec:detailed_results}
\begin{figure}[htbp]
    \centering
    \includegraphics[width=\textwidth]{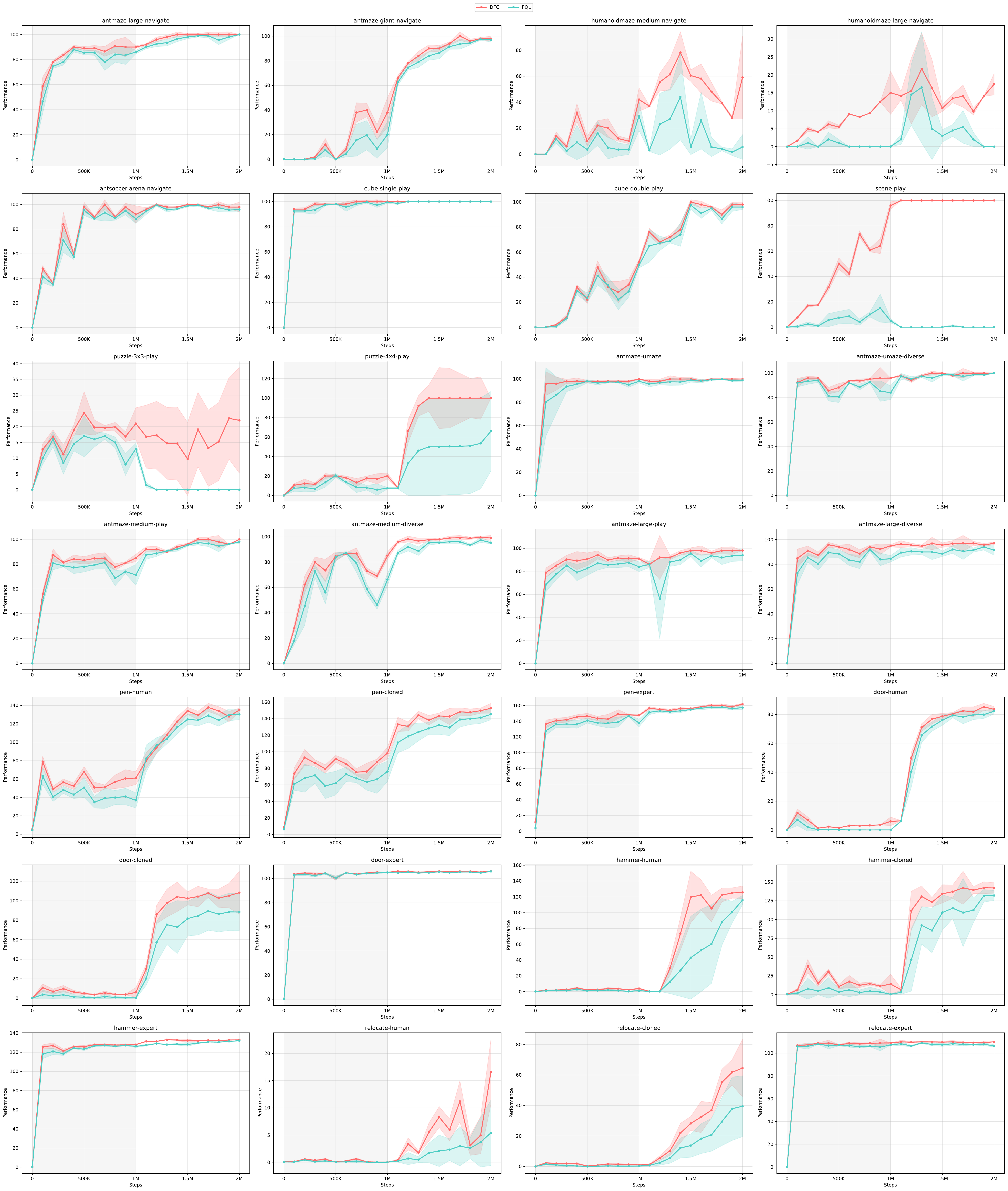}
    \caption{\textbf{Learning curves for offline-to-online RL.} Conducted on the D4RL and OGBench benchmark suites. The initial one million steps, conducted in the offline setting, are highlighted by the gray shaded area.
}
    \label{fig:curve}
    \vspace{-10pt}
\end{figure}

Results on default tasks of OGBench and D4RL tasks are shown in Figure \ref{fig:curve}. Our full results are shown in Table \ref{table:offline_full} and Table \ref{table:overall}.

\begin{table}[h]
\caption{
\footnotesize
\textbf{Offline RL results. }
The following tables present a comprehensive evaluation of our method against key baselines across a suite of over 70 tasks from the OGBench and D4RL benchmarks. To ensure statistical robustness, all experiments were conducted with multiple random seeds: $8$ for state-based tasks and $4$ for pixel-based tasks. The best scores are highlighted in \textbf{bold} and the suboptimal scores are labeled as \underline{underlined}. The asterisk $(*)$ denotes the default task within each environment group.
}
\label{table:offline_full}
\centering
\scalebox{0.6}{ 
\begin{threeparttable}
\begin{tabular}{lccccccc} 
\toprule
\texttt{Task} & \texttt{IQL} & \texttt{ReBRAC} & \texttt{IDQL} & \texttt{CAC} & \texttt{IFQL} & \texttt{FQL} & \texttt{DFC} \\
\midrule
\texttt{antmaze-large-navigate-singletask-task1-v0 (*)} & $48$ {\tiny $\pm 9$} & $\mathbf{91}$ {\tiny $\pm 10$} & $0$ {\tiny $\pm 0$} & $42$ {\tiny $\pm 7$} & $24$ {\tiny $\pm 17$} & $80$ {\tiny $\pm 8$} & $\underline{90}$ {\tiny $\pm 1$} \\
\texttt{antmaze-large-navigate-singletask-task2-v0} & $42$ {\tiny $\pm 6$} & $\mathbf{88}$ {\tiny $\pm 4$} & $14$ {\tiny $\pm 8$} & $1$ {\tiny $\pm 1$} & $8$ {\tiny $\pm 3$} & $57$ {\tiny $\pm 10$} & $\underline{74}$ {\tiny $\pm 4$} \\
\texttt{antmaze-large-navigate-singletask-task3-v0} & $72$ {\tiny $\pm 7$} & $51$ {\tiny $\pm 18$} & $26$ {\tiny $\pm 8$} & $49$ {\tiny $\pm 10$} & $52$ {\tiny $\pm 17$} & $\underline{93}$ {\tiny $\pm 3$} & $\mathbf{98}$ {\tiny $\pm 1$} \\
\texttt{antmaze-large-navigate-singletask-task4-v0} & $51$ {\tiny $\pm 9$} & $\underline{84}$ {\tiny $\pm 7$} & $62$ {\tiny $\pm 25$} & $17$ {\tiny $\pm 6$} & $18$ {\tiny $\pm 8$} & ${80}$ {\tiny $\pm 4$} & $\mathbf{89}$ {\tiny $\pm 2$} \\
\texttt{antmaze-large-navigate-singletask-task5-v0} & $54$ {\tiny $\pm 22$} & $\underline{90}$ {\tiny $\pm 2$} & $2$ {\tiny $\pm 2$} & $55$ {\tiny $\pm 6$} & $38$ {\tiny $\pm 18$} & $83$ {\tiny $\pm 4$} & $\mathbf{91}$ {\tiny $\pm 3$} \\
\midrule
\texttt{antmaze-giant-navigate-singletask-task1-v0 (*)} & $0$ {\tiny $\pm 0$} & $\mathbf{27}$ {\tiny $\pm 22$} & $0$ {\tiny $\pm 0$} & $0$ {\tiny $\pm 0$} & $0$ {\tiny $\pm 0$} & $4$ {\tiny $\pm 5$} & $\underline{25}$ {\tiny $\pm 12$} \\
\texttt{antmaze-giant-navigate-singletask-task2-v0} & $1$ {\tiny $\pm 1$} & $\mathbf{16}$ {\tiny $\pm 17$} & $0$ {\tiny $\pm 0$} & $0$ {\tiny $\pm 0$} & $0$ {\tiny $\pm 0$} & $\underline{9}$ {\tiny $\pm 7$} & $1$ {\tiny $\pm 1$} \\
\texttt{antmaze-giant-navigate-singletask-task3-v0} & $0$ {\tiny $\pm 0$} & $\mathbf{34}$ {\tiny $\pm 22$} & $0$ {\tiny $\pm 0$} & $0$ {\tiny $\pm 0$} & $0$ {\tiny $\pm 0$} & $0$ {\tiny $\pm 1$} & $\underline{1}$ {\tiny $\pm 1$} \\
\texttt{antmaze-giant-navigate-singletask-task4-v0} & $0$ {\tiny $\pm 0$} & $5$ {\tiny $\pm 12$} & $0$ {\tiny $\pm 0$} & $0$ {\tiny $\pm 0$} & $0$ {\tiny $\pm 0$} & $\underline{14}$ {\tiny $\pm 23$} & $\mathbf{40}$ {\tiny $\pm 45$} \\
\texttt{antmaze-giant-navigate-singletask-task5-v0} & ${19}$ {\tiny $\pm 7$} & $\mathbf{49}$ {\tiny $\pm 22$} & $0$ {\tiny $\pm 1$} & $0$ {\tiny $\pm 0$} & $13$ {\tiny $\pm 9$} & $16$ {\tiny $\pm 28$} & $\underline{30}$ {\tiny $\pm 10$} \\
\midrule
\texttt{humanoidmaze-medium-navigate-singletask-task1-v0 (*)} & $32$ {\tiny $\pm 7$} & $16$ {\tiny $\pm 9$} & $1$ {\tiny $\pm 1$} & $\underline{38}$ {\tiny $\pm 19$} & $\mathbf{69}$ {\tiny $\pm 19$} & $19$ {\tiny $\pm 12$} & ${35}$ {\tiny $\pm 9$} \\
\texttt{humanoidmaze-medium-navigate-singletask-task2-v0} & $41$ {\tiny $\pm 9$} & $18$ {\tiny $\pm 16$} & $1$ {\tiny $\pm 1$} & $47$ {\tiny $\pm 35$} & $85$ {\tiny $\pm 11$} & $\underline{94}$ {\tiny $\pm 3$} & $\mathbf{99}$ {\tiny $\pm 1$} \\
\texttt{humanoidmaze-medium-navigate-singletask-task3-v0} & $25$ {\tiny $\pm 5$} & $36$ {\tiny $\pm 13$} & $0$ {\tiny $\pm 1$} & $\underline{83}$ {\tiny $\pm 18$} & $49$ {\tiny $\pm 49$} & $74$ {\tiny $\pm 18$} & $\mathbf{96}$ {\tiny $\pm 3$} \\
\texttt{humanoidmaze-medium-navigate-singletask-task4-v0} & $0$ {\tiny $\pm 1$} & $\mathbf{15}$ {\tiny $\pm 16$} & $1$ {\tiny $\pm 1$} & $\underline{5}$ {\tiny $\pm 4$} & $1$ {\tiny $\pm 1$} & $3$ {\tiny $\pm 4$} & $3$ {\tiny $\pm 1$} \\
\texttt{humanoidmaze-medium-navigate-singletask-task5-v0} & $66$ {\tiny $\pm 4$} & $24$ {\tiny $\pm 20$} & $1$ {\tiny $\pm 1$} & $91$ {\tiny $\pm 5$} & $\underline{98}$ {\tiny $\pm 2$} & ${97}$ {\tiny $\pm 2$} & $\mathbf{100}$ {\tiny $\pm 1$} \\
\midrule
\texttt{humanoidmaze-large-navigate-singletask-task1-v0 (*)} & $3$ {\tiny $\pm 1$} & $2$ {\tiny $\pm 1$} & $0$ {\tiny $\pm 0$} & $1$ {\tiny $\pm 1$} & $6$ {\tiny $\pm 2$} & $\underline{7}$ {\tiny $\pm 6$} & $\mathbf{15}$ {\tiny $\pm 6$} \\
\texttt{humanoidmaze-large-navigate-singletask-task2-v0} & $\mathbf{0}$ {\tiny $\pm 0$} & $\mathbf{0}$ {\tiny $\pm 0$} & $\mathbf{0}$ {\tiny $\pm 0$} & $\mathbf{0}$ {\tiny $\pm 0$} & $\mathbf{0}$ {\tiny $\pm 0$} & $\mathbf{0}$ {\tiny $\pm 0$} & $\mathbf{0}$ {\tiny $\pm 0$} \\
\texttt{humanoidmaze-large-navigate-singletask-task3-v0} & $7$ {\tiny $\pm 3$} & $8$ {\tiny $\pm 4$} & $3$ {\tiny $\pm 1$} & $2$ {\tiny $\pm 3$} & $\mathbf{48}$ {\tiny $\pm 10$} & $11$ {\tiny $\pm 7$} & $\underline{19}$ {\tiny $\pm 5$} \\
\texttt{humanoidmaze-large-navigate-singletask-task4-v0} & $1$ {\tiny $\pm 0$} & $1$ {\tiny $\pm 1$} & $0$ {\tiny $\pm 0$} & $0$ {\tiny $\pm 1$} & $1$ {\tiny $\pm 1$} & $\underline{2}$ {\tiny $\pm 3$} & $\mathbf{4}$ {\tiny $\pm 2$} \\
\texttt{humanoidmaze-large-navigate-singletask-task5-v0} & $1$ {\tiny $\pm 1$} & $\mathbf{2}$ {\tiny $\pm 2$} & $0$ {\tiny $\pm 0$} & $0$ {\tiny $\pm 0$} & $0$ {\tiny $\pm 0$} & $1$ {\tiny $\pm 3$} & $\underline{1}$ {\tiny $\pm 2$} \\
\midrule
\texttt{antsoccer-arena-navigate-singletask-task1-v0} & $14$ {\tiny $\pm 5$} & $0$ {\tiny $\pm 0$} & $44$ {\tiny $\pm 12$} & $1$ {\tiny $\pm 3$} & $61$ {\tiny $\pm 25$} & $\underline{77}$ {\tiny $\pm 4$} & $\mathbf{84}$ {\tiny $\pm 3$} \\
\texttt{antsoccer-arena-navigate-singletask-task2-v0} & $17$ {\tiny $\pm 7$} & $0$ {\tiny $\pm 1$} & $15$ {\tiny $\pm 12$} & $0$ {\tiny $\pm 0$} & $75$ {\tiny $\pm 3$} & $\underline{88}$ {\tiny $\pm 3$} & $\mathbf{96}$ {\tiny $\pm 3$} \\
\texttt{antsoccer-arena-navigate-singletask-task3-v0} & $6$ {\tiny $\pm 4$} & $0$ {\tiny $\pm 0$} & $0$ {\tiny $\pm 0$} & $8$ {\tiny $\pm 19$} & $14$ {\tiny $\pm 22$} & $\underline{61}$ {\tiny $\pm 6$} & $\underline{69}$ {\tiny $\pm 5$} \\
\texttt{antsoccer-arena-navigate-singletask-task4-v0 (*)} & $3$ {\tiny $\pm 2$} & $0$ {\tiny $\pm 0$} & $0$ {\tiny $\pm 1$} & $0$ {\tiny $\pm 0$} & $16$ {\tiny $\pm 9$} & $\underline{39}$ {\tiny $\pm 6$} & $\mathbf{52}$ {\tiny $\pm 7$} \\
\texttt{antsoccer-arena-navigate-singletask-task5-v0} & $2$ {\tiny $\pm 2$} & $0$ {\tiny $\pm 0$} & $0$ {\tiny $\pm 0$} & $0$ {\tiny $\pm 0$} & $0$ {\tiny $\pm 1$} & $\underline{36}$ {\tiny $\pm 9$} & $\mathbf{66}$ {\tiny $\pm 5$} \\
\midrule
\texttt{cube-single-play-singletask-task1-v0} & $88$ {\tiny $\pm 3$} & $89$ {\tiny $\pm 5$} & $\mathbf{95}$ {\tiny $\pm 2$} & $77$ {\tiny $\pm 28$} & $79$ {\tiny $\pm 4$} & $\underline{97}$ {\tiny $\pm 2$} & $\mathbf{100}$ {\tiny $\pm 0$} \\
\texttt{cube-single-play-singletask-task2-v0 (*)} & $85$ {\tiny $\pm 8$} & $92$ {\tiny $\pm 4$} & $\mathbf{96}$ {\tiny $\pm 2$} & $80$ {\tiny $\pm 30$} & $73$ {\tiny $\pm 3$} & $\underline{97}$ {\tiny $\pm 2$} & $\mathbf{100}$ {\tiny $\pm 0$} \\
\texttt{cube-single-play-singletask-task3-v0} & $91$ {\tiny $\pm 5$} & $93$ {\tiny $\pm 3$} & $\mathbf{99}$ {\tiny $\pm 1$} & $\mathbf{98}$ {\tiny $\pm 1$} & $88$ {\tiny $\pm 4$} & $\underline{98}$ {\tiny $\pm 2$} & $\mathbf{100}$ {\tiny $\pm 0$} \\
\texttt{cube-single-play-singletask-task4-v0} & $73$ {\tiny $\pm 6$} & ${92}$ {\tiny $\pm 3$} & ${93}$ {\tiny $\pm 4$} & ${91}$ {\tiny $\pm 2$} & $79$ {\tiny $\pm 6$} & $\underline{94}$ {\tiny $\pm 3$} & $\mathbf{99}$ {\tiny $\pm 1$} \\
\texttt{cube-single-play-singletask-task5-v0} & $78$ {\tiny $\pm 9$} & $87$ {\tiny $\pm 8$} & ${90}$ {\tiny $\pm 6$} & $80$ {\tiny $\pm 20$} & $77$ {\tiny $\pm 7$} & $\underline{93}$ {\tiny $\pm 3$} & $\mathbf{100}$ {\tiny $\pm 0$} \\
\midrule
\texttt{cube-double-play-singletask-task1-v0} & $27$ {\tiny $\pm 5$} & $45$ {\tiny $\pm 6$} & $39$ {\tiny $\pm 19$} & $21$ {\tiny $\pm 8$} & $35$ {\tiny $\pm 9$} & $\underline{61}$ {\tiny $\pm 9$} & $\mathbf{84}$ {\tiny $\pm 4$} \\
\texttt{cube-double-play-singletask-task2-v0 (*)} & $1$ {\tiny $\pm 1$} & $7$ {\tiny $\pm 3$} & $16$ {\tiny $\pm 10$} & $2$ {\tiny $\pm 2$} & $9$ {\tiny $\pm 5$} & $\underline{36}$ {\tiny $\pm 6$} & $\mathbf{49}$ {\tiny $\pm 2$} \\
\texttt{cube-double-play-singletask-task3-v0} & $0$ {\tiny $\pm 0$} & $4$ {\tiny $\pm 1$} & $17$ {\tiny $\pm 8$} & $3$ {\tiny $\pm 1$} & $8$ {\tiny $\pm 5$} & $\underline{22}$ {\tiny $\pm 5$} & $\mathbf{31}$ {\tiny $\pm 1$} \\
\texttt{cube-double-play-singletask-task4-v0} & $0$ {\tiny $\pm 0$} & $1$ {\tiny $\pm 1$} & $0$ {\tiny $\pm 1$} & $0$ {\tiny $\pm 1$} & $1$ {\tiny $\pm 1$} & $\underline{5}$ {\tiny $\pm 2$} & $\mathbf{11}$ {\tiny $\pm 4$} \\
\texttt{cube-double-play-singletask-task5-v0} & $4$ {\tiny $\pm 3$} & $4$ {\tiny $\pm 2$} & $1$ {\tiny $\pm 1$} & $3$ {\tiny $\pm 2$} & $17$ {\tiny $\pm 6$} & $\underline{19}$ {\tiny $\pm 10$} & $\mathbf{28}$ {\tiny $\pm 11$} \\
\midrule
\texttt{scene-play-singletask-task1-v0} & $94$ {\tiny $\pm 3$} & ${95}$ {\tiny $\pm 2$} & $\mathbf{100}$ {\tiny $\pm 0$} & $\mathbf{100}$ {\tiny $\pm 1$} & ${98}$ {\tiny $\pm 3$} & $\mathbf{100}$ {\tiny $\pm 0$} & $\mathbf{100}$ {\tiny $\pm 0$} \\
\texttt{scene-play-singletask-task2-v0 (*)} & $12$ {\tiny $\pm 3$} & $50$ {\tiny $\pm 13$} & $33$ {\tiny $\pm 14$} & $50$ {\tiny $\pm 40$} & $0$ {\tiny $\pm 0$} & $\underline{76}$ {\tiny $\pm 9$} & $\mathbf{96}$ {\tiny $\pm 3$} \\
\texttt{scene-play-singletask-task3-v0} & $32$ {\tiny $\pm 7$} & $55$ {\tiny $\pm 16$} & ${94}$ {\tiny $\pm 4$} & $49$ {\tiny $\pm 16$} & $54$ {\tiny $\pm 19$} & $\underline{98}$ {\tiny $\pm 1$} & $\mathbf{100}$ {\tiny $\pm 0$} \\
\texttt{scene-play-singletask-task4-v0} & $0$ {\tiny $\pm 1$} & $3$ {\tiny $\pm 3$} & $4$ {\tiny $\pm 3$} & $0$ {\tiny $\pm 0$} & $0$ {\tiny $\pm 0$} & $\underline{5}$ {\tiny $\pm 1$} & $\mathbf{20}$ {\tiny $\pm 9$} \\
\texttt{scene-play-singletask-task5-v0} & $\mathbf{0}$ {\tiny $\pm 0$} & $\mathbf{0}$ {\tiny $\pm 0$} & $\mathbf{0}$ {\tiny $\pm 0$} & $\mathbf{0}$ {\tiny $\pm 0$} & $\mathbf{0}$ {\tiny $\pm 0$} & $\mathbf{0}$ {\tiny $\pm 0$} & $\mathbf{0}$ {\tiny $\pm 0$} \\
\midrule
\texttt{puzzle-3x3-play-singletask-task1-v0} & $33$ {\tiny $\pm 6$} & $\mathbf{97}$ {\tiny $\pm 4$} & $52$ {\tiny $\pm 12$} & $\mathbf{97}$ {\tiny $\pm 2$} & $\underline{94}$ {\tiny $\pm 3$} & $90$ {\tiny $\pm 4$} & $\mathbf{97}$ {\tiny $\pm 1$} \\
\texttt{puzzle-3x3-play-singletask-task2-v0} & $4$ {\tiny $\pm 3$} & $1$ {\tiny $\pm 1$} & $0$ {\tiny $\pm 1$} & $0$ {\tiny $\pm 0$} & $1$ {\tiny $\pm 2$} & $\underline{16}$ {\tiny $\pm 5$} & $\mathbf{36}$ {\tiny $\pm 1$} \\
\texttt{puzzle-3x3-play-singletask-task3-v0} & $3$ {\tiny $\pm 2$} & $3$ {\tiny $\pm 1$} & $0$ {\tiny $\pm 0$} & $0$ {\tiny $\pm 0$} & $0$ {\tiny $\pm 0$} & $\underline{10}$ {\tiny $\pm 3$} & $\mathbf{17}$ {\tiny $\pm 3$} \\
\texttt{puzzle-3x3-play-singletask-task4-v0 (*)} & $2$ {\tiny $\pm 1$} & $2$ {\tiny $\pm 1$} & $0$ {\tiny $\pm 0$} & $0$ {\tiny $\pm 0$} & $0$ {\tiny $\pm 0$} & $\mathbf{16}$ {\tiny $\pm 5$} & $\mathbf{21}$ {\tiny $\pm 5$} \\
\texttt{puzzle-3x3-play-singletask-task5-v0} & $3$ {\tiny $\pm 2$} & $5$ {\tiny $\pm 3$} & $0$ {\tiny $\pm 0$} & $0$ {\tiny $\pm 0$} & $0$ {\tiny $\pm 0$} & $\underline{16}$ {\tiny $\pm 3$} & $\mathbf{29}$ {\tiny $\pm 5$} \\
\midrule
\texttt{puzzle-4x4-play-singletask-task1-v0} & $12$ {\tiny $\pm 2$} & $26$ {\tiny $\pm 4$} & ${48}$ {\tiny $\pm 5$} & $44$ {\tiny $\pm 10$} & $\underline{49}$ {\tiny $\pm 9$} & $34$ {\tiny $\pm 8$} & $\mathbf{68}$ {\tiny $\pm 5$} \\
\texttt{puzzle-4x4-play-singletask-task2-v0} & $7$ {\tiny $\pm 4$} & $12$ {\tiny $\pm 4$} & $14$ {\tiny $\pm 5$} & $0$ {\tiny $\pm 0$} & $4$ {\tiny $\pm 4$} & $\underline{16}$ {\tiny $\pm 5$} & $\mathbf{24}$ {\tiny $\pm 2$} \\
\texttt{puzzle-4x4-play-singletask-task3-v0} & $9$ {\tiny $\pm 3$} & $15$ {\tiny $\pm 3$} & $34$ {\tiny $\pm 5$} & $29$ {\tiny $\pm 12$} & $\mathbf{50}$ {\tiny $\pm 14$} & $18$ {\tiny $\pm 5$} & $\underline{49}$ {\tiny $\pm 3$} \\
\texttt{puzzle-4x4-play-singletask-task4-v0 (*)} & $5$ {\tiny $\pm 2$} & $10$ {\tiny $\pm 3$} & $\mathbf{26}$ {\tiny $\pm 6$} & $1$ {\tiny $\pm 1$} & $21$ {\tiny $\pm 11$} & $11$ {\tiny $\pm 3$} & $\mathbf{20}$ {\tiny $\pm 3$} \\
\texttt{puzzle-4x4-play-singletask-task5-v0} & $4$ {\tiny $\pm 1$} & $7$ {\tiny $\pm 3$} & $\mathbf{24}$ {\tiny $\pm 11$} & $0$ {\tiny $\pm 0$} & $2$ {\tiny $\pm 2$} & $7$ {\tiny $\pm 3$} & $\underline{14}$ {\tiny $\pm 4$} \\
\midrule
\texttt{antmaze-umaze-v2} & $77$ & $\underline{98}$ & ${94}$ & $66$ {\tiny $\pm 5$} & $92$ {\tiny $\pm 6$} & ${96}$ {\tiny $\pm 2$} & $\mathbf{100}$ {\tiny $\pm 0$} \\
\texttt{antmaze-umaze-diverse-v2} & $54$ & $84$ & $80$ & $66$ {\tiny $\pm 11$} & $62$ {\tiny $\pm 12$} & $\underline{89}$ {\tiny $\pm 5$} & $\mathbf{96}$ {\tiny $\pm 0$} \\
\texttt{antmaze-medium-play-v2} & $66$ & $\mathbf{90}$ & $84$ & $49$ {\tiny $\pm 24$} & $56$ {\tiny $\pm 15$} & $78$ {\tiny $\pm 7$} & $\underline{85}$ {\tiny $\pm 3$} \\
\texttt{antmaze-medium-diverse-v2} & ${74}$ & $\underline{84}$ & $\mathbf{85}$ & $0$ {\tiny $\pm 1$} & $60$ {\tiny $\pm 25$} & $71$ {\tiny $\pm 13$} & $\mathbf{85}$ {\tiny $\pm 3$} \\
\texttt{antmaze-large-play-v2} & $42$ & $52$ & $64$ & $0$ {\tiny $\pm 0$} & $55$ {\tiny $\pm 9$} & $\underline{84}$ {\tiny $\pm 7$} & $\mathbf{91}$ {\tiny $\pm 1$} \\
\texttt{antmaze-large-diverse-v2} & $30$ & $64$ & $68$ & $0$ {\tiny $\pm 0$} & $64$ {\tiny $\pm 8$} & $\underline{83}$ {\tiny $\pm 4$} & $\mathbf{95}$ {\tiny $\pm 1$} \\
\midrule
\texttt{pen-human-v1} & $78$ & $\mathbf{103}$ & $\underline{76}$ {\tiny $\pm 10$} & $64$ {\tiny $\pm 8$} & $71$ {\tiny $\pm 12$} & $53$ {\tiny $\pm 6$} & ${61}$ {\tiny $\pm 7$} \\
\texttt{pen-cloned-v1} & $83$ & $\mathbf{103}$ & $64$ {\tiny $\pm 7$} & $56$ {\tiny $\pm 10$} & $80$ {\tiny $\pm 11$} & $74$ {\tiny $\pm 11$} & $\underline{90}$ {\tiny $\pm 7$} \\
\texttt{pen-expert-v1} & $128$ & $\mathbf{152}$ & $140$ {\tiny $\pm 6$} & $103$ {\tiny $\pm 9$} & $139$ {\tiny $\pm 5$} & $142$ {\tiny $\pm 6$} & $\underline{147}$ {\tiny $\pm 1$} \\
\texttt{door-human-v1} & $3$ & $-0$ & $6$ {\tiny $\pm 2$} & $5$ {\tiny $\pm 2$} & $\mathbf{7}$ {\tiny $\pm 2$} & $0$ {\tiny $\pm 0$} & $\underline{6}$ {\tiny $\pm 3$} \\
\texttt{door-cloned-v1} & $\underline{3}$ & $0$ & $0$ {\tiny $\pm 0$} & $1$ {\tiny $\pm 0$} & $2$ {\tiny $\pm 2$} & $2$ {\tiny $\pm 1$} & $\mathbf{6}$ {\tiny $\pm 5$} \\
\texttt{door-expert-v1} & $\mathbf{107}$ & $\underline{106}$ & ${105}$ {\tiny $\pm 1$} & $98$ {\tiny $\pm 3$} & ${104}$ {\tiny $\pm 2$} & ${104}$ {\tiny $\pm 1$} & ${105}$ {\tiny $\pm 0$} \\
\texttt{hammer-human-v1} & $2$ & $0$ & $2$ {\tiny $\pm 1$} & $2$ {\tiny $\pm 0$} & $\mathbf{3}$ {\tiny $\pm 1$} & $1$ {\tiny $\pm 1$} & $\mathbf{4}$ {\tiny $\pm 1$} \\
\texttt{hammer-cloned-v1} & $2$ & $5$ & $2$ {\tiny $\pm 1$} & $1$ {\tiny $\pm 1$} & $2$ {\tiny $\pm 1$} & $\underline{11}$ {\tiny $\pm 9$} & $\mathbf{14}$ {\tiny $\pm 13$} \\
\texttt{hammer-expert-v1} & $\underline{129}$ & $\mathbf{134}$ & $125$ {\tiny $\pm 4$} & $92$ {\tiny $\pm 11$} & $117$ {\tiny $\pm 9$} & $125$ {\tiny $\pm 3$} & ${128}$ {\tiny $\pm 0$} \\
\texttt{relocate-human-v1} & $\mathbf{0}$ & $\mathbf{0}$ & $\mathbf{0}$ {\tiny $\pm 0$} & $\mathbf{0}$ {\tiny $\pm 0$} & $\mathbf{0}$ {\tiny $\pm 0$} & $\mathbf{0}$ {\tiny $\pm 0$} & $\mathbb{0}$ {\tiny $\pm 0$} \\
\texttt{relocate-cloned-v1} & $0$ & $\mathbf{2}$ & $-0$ {\tiny $\pm 0$} & $-0$ {\tiny $\pm 0$} & $-0$ {\tiny $\pm 0$} & $-0$ {\tiny $\pm 0$} & $\underline{1}$ {\tiny $\pm 0$} \\
\texttt{relocate-expert-v1} & ${106}$ & $\underline{108}$ & ${107}$ {\tiny $\pm 1$} & $93$ {\tiny $\pm 6$} & ${104}$ {\tiny $\pm 3$} & ${107}$ {\tiny $\pm 1$} & $\mathbf{109}$ {\tiny $\pm 0$} \\
\midrule
\texttt{visual-cube-single-play-singletask-task1-v0} & $70$ {\tiny $\pm 12$} & $\underline{83}$ {\tiny $\pm 6$} & - & - & $49$ {\tiny $\pm 7$} & ${81}$ {\tiny $\pm 12$} & $\mathbf{85}$ {\tiny $\pm 10$} \\
\texttt{visual-cube-double-play-singletask-task1-v0} & $\mathbf{34}$ {\tiny $\pm 23$} & $4$ {\tiny $\pm 4$} & - & - & $8$ {\tiny $\pm 6$} & $21$ {\tiny $\pm 11$} & $\mathbf{23}$ {\tiny $\pm 10$} \\
\texttt{visual-scene-play-singletask-task1-v0} & ${97}$ {\tiny $\pm 2$} & $\underline{98}$ {\tiny $\pm 4$} & - & - & $86$ {\tiny $\pm 10$} & $\underline{98}$ {\tiny $\pm 3$} & $\mathbf{100}$ {\tiny $\pm 0$} \\
\texttt{visual-puzzle-3x3-play-singletask-task1-v0} & $7$ {\tiny $\pm 15$} & $88$ {\tiny $\pm 4$} & - & - & $\mathbf{100}$ {\tiny $\pm 0$} & $94$ {\tiny $\pm 1$} & $\underline{95}$ {\tiny $\pm 7$} \\
\texttt{visual-puzzle-4x4-play-singletask-task1-v0} & $0$ {\tiny $\pm 0$} & $26$ {\tiny $\pm 6$} & - & - & $8$ {\tiny $\pm 15$} & $\underline{33}$ {\tiny $\pm 6$} & $\mathbf{37}$ {\tiny $\pm 6$} \\
\bottomrule
\end{tabular}
\end{threeparttable}
}
\vspace{-10pt}
\end{table}

\begin{table}[h]
\caption{
\footnotesize
\textbf{Ablation study of the DFC model on a single-layer critic architecture.}
 We compare the full model against its key components: a Flow-only Critic (FC) and a Distributional-only Critic (DC). FQL is included for baseline performance reference. 
 The asterisk $(*)$ denotes the default task within each environment group.
}
\label{table:overall}
\centering
\scalebox{0.6}{ 
\begin{threeparttable}
\begin{tabular}{lccccc} 
\toprule
\texttt{Task} & \texttt{FQL} & \texttt{FC} & \texttt{DC} & \texttt{DFC} \\
\midrule
\texttt{antmaze-large-navigate-singletask-task1-v0 (*)} & $80$ {\tiny $\pm 8$} $\to$ $100$ {\tiny $\pm 0$} & $86$ {\tiny $\pm 3$} $\to$ $100$ {\tiny $\pm 0$} & $88$ {\tiny $\pm 1$} $\to$ $100$ {\tiny $\pm 0$} & ${90}$ {\tiny $\pm 1$} $\to$ $100$ {\tiny $\pm 0$} \\
\texttt{antmaze-large-navigate-singletask-task2-v0} & $57$ {\tiny $\pm 10$} $\to$ $88$ {\tiny $\pm 2$} & $72$ {\tiny $\pm 16$} $\to$ $88$ {\tiny $\pm 3$} & $69$ {\tiny $\pm 8$} $\to$ $88$ {\tiny $\pm 1$} & ${74}$ {\tiny $\pm 4$}  $\to$ $89$ {\tiny $\pm 1$} \\
\texttt{antmaze-large-navigate-singletask-task3-v0} & ${93}$ {\tiny $\pm 3$} $\to$ $100$ {\tiny $\pm 0$} & $96$ {\tiny $\pm 6$} $\to$ $100$ {\tiny $\pm 0$} & $96$ {\tiny $\pm 2$} $\to$ $100$ {\tiny $\pm 0$} & ${98}$ {\tiny $\pm 1$}  $\to$ $100$ {\tiny $\pm 0$} \\
\texttt{antmaze-large-navigate-singletask-task4-v0} & ${80}$ {\tiny $\pm 4$} $\to$ $97$ {\tiny $\pm 1$} & $86$ {\tiny $\pm 4$} $\to$ $98$ {\tiny $\pm 1$} & $88$ {\tiny $\pm 2$} $\to$ $98$ {\tiny $\pm 0$} & ${89}$ {\tiny $\pm 2$}  $\to$ $99$ {\tiny $\pm 1$} \\
\texttt{antmaze-large-navigate-singletask-task5-v0} & $83$ {\tiny $\pm 4$} $\to$ $99$ {\tiny $\pm 1$} & $90$ {\tiny $\pm 6$} $\to$ $100$ {\tiny $\pm 1$} & $91$ {\tiny $\pm 3$} $\to$ $100$ {\tiny $\pm 1$} & ${91}$ {\tiny $\pm 3$}  $\to$ $100$ {\tiny $\pm 1$} \\
\midrule
\texttt{antmaze-giant-navigate-singletask-task1-v0 (*)} & $4$ {\tiny $\pm 5$} $\to$ $97$ {\tiny $\pm 2$} & $16$ {\tiny $\pm 21$} $\to$ $98$ {\tiny $\pm 3$} & $22$ {\tiny $\pm 12$} $\to$ $94$ {\tiny $\pm 1$} & ${25}$ {\tiny $\pm 12$}  $\to$ $98$ {\tiny $\pm 2$} \\
\texttt{antmaze-giant-navigate-singletask-task2-v0} & $9$ {\tiny $\pm 7$} $\to$ $98$ {\tiny $\pm 0$} & $0$ {\tiny $\pm 1$} $\to$ $98$ {\tiny $\pm 0$} & $1$ {\tiny $\pm 1$} $\to$ $100$ {\tiny $\pm 0$} & $1$ {\tiny $\pm 1$} $\to$ $99$ {\tiny $\pm 1$} \\
\texttt{antmaze-giant-navigate-singletask-task3-v0} & $0$ {\tiny $\pm 1$} $\to$ $0$ {\tiny $\pm 0$} & $0$ {\tiny $\pm 0$} $\to$ $0$ {\tiny $\pm 0$} & $1$ {\tiny $\pm 0$} $\to$ $3$ {\tiny $\pm 4$} & $1$ {\tiny $\pm 1$} $\to$ $10$ {\tiny $\pm 3$} \\
\texttt{antmaze-giant-navigate-singletask-task4-v0} & ${14}$ {\tiny $\pm 23$} $\to$ $95$ {\tiny $\pm 1$} & $38$ {\tiny $\pm 54$} $\to$ $94$ {\tiny $\pm 1$} & $39$ {\tiny $\pm 44$} $\to$ $94$ {\tiny $\pm 1$} & ${40}$ {\tiny $\pm 45$}  $\to$ $99$ {\tiny $\pm 1$} \\
\texttt{antmaze-giant-navigate-singletask-task5-v0} & $16$ {\tiny $\pm 28$} $\to$ $100$ {\tiny $\pm 0$} & $0$ {\tiny $\pm 6$} $\to$ $100$ {\tiny $\pm 0$} & $16$ {\tiny $\pm 7$} $\to$ $100$ {\tiny $\pm 0$} & ${30}$ {\tiny $\pm 10$} $\to$ $100$ {\tiny $\pm 0$} \\
\midrule
\texttt{humanoidmaze-medium-navigate-singletask-task1-v0 (*)} & $19$ {\tiny $\pm 12$} $\to$ $22$ {\tiny $\pm 13$} & $40$ {\tiny $\pm 17$} $\to$ $0$ {\tiny $\pm 16$} & $34$ {\tiny $\pm 9$} $\to$ $34$ {\tiny $\pm 11$} & ${35}$ {\tiny $\pm 9$} $\to$ $59$ {\tiny $\pm 32$} \\
\texttt{humanoidmaze-medium-navigate-singletask-task2-v0} & ${94}$ {\tiny $\pm 3$} $\to$ $99$ {\tiny $\pm 1$} & $98$ {\tiny $\pm 3$} $\to$ $98$ {\tiny $\pm 1$} & $98$ {\tiny $\pm 1$} $\to$ $98$ {\tiny $\pm 1$} & ${99}$ {\tiny $\pm 1$} $\to$ $100$ {\tiny $\pm 0$} \\
\texttt{humanoidmaze-medium-navigate-singletask-task3-v0} & $74$ {\tiny $\pm 18$} $\to$ $79$ {\tiny $\pm 16$} & $98$ {\tiny $\pm 24$} $\to$ $0$ {\tiny $\pm 20$} & $90$ {\tiny $\pm 10$} $\to$ $90$ {\tiny $\pm 49$} & ${96}$ {\tiny $\pm 3$} $\to$ $99$ {\tiny $\pm 1$} \\
\texttt{humanoidmaze-medium-navigate-singletask-task4-v0} & $3$ {\tiny $\pm 4$} $\to$ $13$ {\tiny $\pm 19$} & $2$ {\tiny $\pm 1$} $\to$ $4$ {\tiny $\pm 24$} & $2$ {\tiny $\pm 1$} $\to$ $34$ {\tiny $\pm 3$} & $3$ {\tiny $\pm 1$} $\to$ $66$ {\tiny $\pm 24$} \\
\texttt{humanoidmaze-medium-navigate-singletask-task5-v0} & ${97}$ {\tiny $\pm 2$} $\to$ $77$ {\tiny $\pm 40$} & $100$ {\tiny $\pm 3$} $\to$ $100$ {\tiny $\pm 49$} & $99$ {\tiny $\pm 1$} $\to$ $99$ {\tiny $\pm 0$} & ${100}$ {\tiny $\pm 1$} $\to$ $100$ {\tiny $\pm 0$} \\
\midrule
\texttt{humanoidmaze-large-navigate-singletask-task1-v0 (*)} & ${7}$ {\tiny $\pm 6$} $\to$ $0$ {\tiny $\pm 0$} & $0$ {\tiny $\pm 6$} $\to$ $0$ {\tiny $\pm 0$} & $9$ {\tiny $\pm 5$} $\to$ $9$ {\tiny $\pm 10$} & ${15}$ {\tiny $\pm 6$} $\to$ $16$ {\tiny $\pm 3$} \\
\texttt{humanoidmaze-large-navigate-singletask-task2-v0} & ${0}$ {\tiny $\pm 0$} $\to$ $0$ {\tiny $\pm 0$} & $0$ {\tiny $\pm 0$} $\to$ $0$ {\tiny $\pm 0$} & $0$ {\tiny $\pm 0$} $\to$ $0$ {\tiny $\pm 0$} & $0$ {\tiny $\pm 0$} $\to$ $0$ {\tiny $\pm 0$} \\
\texttt{humanoidmaze-large-navigate-singletask-task3-v0} & $11$ {\tiny $\pm 7$} $\to$ $26$ {\tiny $\pm 35$} & $18$ {\tiny $\pm 16$} $\to$ $12$ {\tiny $\pm 47$} & $16$ {\tiny $\pm 7$} $\to$ $32$ {\tiny $\pm 1$} & ${19}$ {\tiny $\pm 5$} $\to$ $67$ {\tiny $\pm 3$} \\
\texttt{humanoidmaze-large-navigate-singletask-task4-v0} & ${2}$ {\tiny $\pm 3$} $\to$ $0$ {\tiny $\pm 0$} & $0$ {\tiny $\pm 0$} $\to$ $0$ {\tiny $\pm 0$} & $2$ {\tiny $\pm 1$} $\to$ $5$ {\tiny $\pm 3$} & ${4}$ {\tiny $\pm 2$} $\to$ $20$ {\tiny $\pm 14$} \\
\texttt{humanoidmaze-large-navigate-singletask-task5-v0} & $1$ {\tiny $\pm 3$} $\to$ $0$ {\tiny $\pm 0$} & $0$ {\tiny $\pm 0$} $\to$ $0$ {\tiny $\pm 0$} & $1$ {\tiny $\pm 1$} $\to$ $1$ {\tiny $\pm 0$} & $1$ {\tiny $\pm 2$} $\to$ $3$ {\tiny $\pm 2$} \\
\midrule
\texttt{antsoccer-arena-navigate-singletask-task1-v0} & ${77}$ {\tiny $\pm 4$} $\to$ $99$ {\tiny $\pm 1$} & $74$ {\tiny $\pm 16$} $\to$ $98$ {\tiny $\pm 1$} & $80$ {\tiny $\pm 6$} $\to$ $100$ {\tiny $\pm 1$} & ${84}$ {\tiny $\pm 3$} $\to$ $100$ {\tiny $\pm 1$} \\
\texttt{antsoccer-arena-navigate-singletask-task2-v0} & ${88}$ {\tiny $\pm 3$} $\to$ $95$ {\tiny $\pm 2$} & $90$ {\tiny $\pm 7$} $\to$ $94$ {\tiny $\pm 3$} & $92$ {\tiny $\pm 3$} $\to$ $94$ {\tiny $\pm 1$} & ${96}$ {\tiny $\pm 3$} $\to$ $100$ {\tiny $\pm 0$} \\
\texttt{antsoccer-arena-navigate-singletask-task3-v0} & ${61}$ {\tiny $\pm 6$} $\to$ $91$ {\tiny $\pm 3$} & $60$ {\tiny $\pm 18$} $\to$ $90$ {\tiny $\pm 4$} & $65$ {\tiny $\pm 8$} $\to$ $90$ {\tiny $\pm 1$} & ${69}$ {\tiny $\pm 5$} $\to$ $94$ {\tiny $\pm 2$} \\
\texttt{antsoccer-arena-navigate-singletask-task4-v0 (*)} & ${39}$ {\tiny $\pm 6$} $\to$ $71$ {\tiny $\pm 4$} & $50$ {\tiny $\pm 21$} $\to$ $70$ {\tiny $\pm 6$} & $50$ {\tiny $\pm 9$} $\to$ $68$ {\tiny $\pm 1$} & ${52}$ {\tiny $\pm 7$} $\to$ $82$ {\tiny $\pm 4$} \\
\texttt{antsoccer-arena-navigate-singletask-task5-v0} & ${36}$ {\tiny $\pm 9$} $\to$ $69$ {\tiny $\pm 5$} & $50$ {\tiny $\pm 8$} $\to$ $72$ {\tiny $\pm 6$} & $57$ {\tiny $\pm 5$} $\to$ $72$ {\tiny $\pm 8$} & ${66}$ {\tiny $\pm 5$} $\to$ $86$ {\tiny $\pm 1$} \\
\midrule
\texttt{cube-single-play-singletask-task1-v0} & ${97}$ {\tiny $\pm 2$} $\to$ $100$ {\tiny $\pm 0$} & $98$ {\tiny $\pm 7$} $\to$ $100$ {\tiny $\pm 0$} & $98$ {\tiny $\pm 2$} $\to$ $100$ {\tiny $\pm 0$} & ${100}$ {\tiny $\pm 0$} $\to$ $100$ {\tiny $\pm 0$} \\
\texttt{cube-single-play-singletask-task2-v0 (*)} & ${97}$ {\tiny $\pm 2$} $\to$ $100$ {\tiny $\pm 0$} & $100$ {\tiny $\pm 1$} $\to$ $100$ {\tiny $\pm 0$} & $100$ {\tiny $\pm 0$} $\to$ $100$ {\tiny $\pm 0$} & ${100}$ {\tiny $\pm 0$} $\to$ $100$ {\tiny $\pm 0$} \\
\texttt{cube-single-play-singletask-task3-v0} & ${98}$ {\tiny $\pm 2$} $\to$ $100$ {\tiny $\pm 0$} & $98$ {\tiny $\pm 1$} $\to$ $100$ {\tiny $\pm 0$} & $100$ {\tiny $\pm 0$} $\to$ $100$ {\tiny $\pm 0$} & ${100}$ {\tiny $\pm 0$} $\to$ $100$ {\tiny $\pm 0$} \\
\texttt{cube-single-play-singletask-task4-v0} & ${94}$ {\tiny $\pm 3$} $\to$ $100$ {\tiny $\pm 0$} & $96$ {\tiny $\pm 4$} $\to$ $100$ {\tiny $\pm 0$} & $97$ {\tiny $\pm 1$} $\to$ $100$ {\tiny $\pm 0$} & ${99}$ {\tiny $\pm 1$} $\to$ $100$ {\tiny $\pm 0$} \\
\texttt{cube-single-play-singletask-task5-v0} & ${93}$ {\tiny $\pm 3$} $\to$ $99$ {\tiny $\pm 1$} & $94$ {\tiny $\pm 4$} $\to$ $100$ {\tiny $\pm 1$} & $97$ {\tiny $\pm 1$} $\to$ $98$ {\tiny $\pm 0$} & ${100}$ {\tiny $\pm 0$} $\to$ $100$ {\tiny $\pm 0$} \\
\midrule
\texttt{cube-double-play-singletask-task1-v0} & ${61}$ {\tiny $\pm 9$} $\to$ $97$ {\tiny $\pm 3$} & $78$ {\tiny $\pm 20$} $\to$ $96$ {\tiny $\pm 4$} & $80$ {\tiny $\pm 8$} $\to$  $96$ {\tiny $\pm 0$} & ${84}$ {\tiny $\pm 4$} $\to$ $100$ {\tiny $\pm 0$} \\
\texttt{cube-double-play-singletask-task2-v0 (*)} & ${36}$ {\tiny $\pm 6$} $\to$ $95$ {\tiny $\pm 2$} & $50$ {\tiny $\pm 6$} $\to$ $94$ {\tiny $\pm 3$} & $49$ {\tiny $\pm 3$} $\to$ $94$ {\tiny $\pm 1$} & ${49}$ {\tiny $\pm 2$} $\to$ $98$ {\tiny $\pm 2$} \\
\texttt{cube-double-play-singletask-task3-v0} & ${22}$ {\tiny $\pm 5$} $\to$ $95$ {\tiny $\pm 3$} & $26$ {\tiny $\pm 7$} $\to$ $94$ {\tiny $\pm 4$} & $29$ {\tiny $\pm 3$} $\to$ $98$ {\tiny $\pm 0$} & ${31}$ {\tiny $\pm 1$} $\to$ $100$ {\tiny $\pm 1$} \\
\texttt{cube-double-play-singletask-task4-v0} & ${5}$ {\tiny $\pm 2$} $\to$ $29$ {\tiny $\pm 23$} & $8$ {\tiny $\pm 4$} $\to$ $32$ {\tiny $\pm 33$} & $9$ {\tiny $\pm 3$} $\to$ $19$ {\tiny $\pm 23$} & ${11}$ {\tiny $\pm 4$} $\to$ $34$ {\tiny $\pm 16$} \\
\texttt{cube-double-play-singletask-task5-v0} & ${19}$ {\tiny $\pm 10$} $\to$ $97$ {\tiny $\pm 4$} & $24$ {\tiny $\pm 18$} $\to$ $98$ {\tiny $\pm 6$} & $28$ {\tiny $\pm 11$} $\to$ $96$ {\tiny $\pm 2$} & ${28}$ {\tiny $\pm 11$} $\to$ $100$ {\tiny $\pm 1$} \\
\midrule
\texttt{scene-play-singletask-task1-v0} & ${100}$ {\tiny $\pm 0$} $\to$ $100$ {\tiny $\pm 0$} & $100$ {\tiny $\pm 1$} $\to$ $100$ {\tiny $\pm 0$} & $100$ {\tiny $\pm 1$} $\to$ $100$ {\tiny $\pm 0$} & ${100}$ {\tiny $\pm 0$} $\to$ $100$ {\tiny $\pm 0$} \\
\texttt{scene-play-singletask-task2-v0 (*)} & ${76}$ {\tiny $\pm 9$} $\to$ $100$ {\tiny $\pm 0$} & $92$ {\tiny $\pm 6$} $\to$ $100$ {\tiny $\pm 0$} & $94$ {\tiny $\pm 4$} $\to$ $96$ {\tiny $\pm 1$} & ${96}$ {\tiny $\pm 3$} $\to$ $100$ {\tiny $\pm 0$} \\
\texttt{scene-play-singletask-task3-v0} & ${98}$ {\tiny $\pm 1$} $\to$ $100$ {\tiny $\pm 0$} & $100$ {\tiny $\pm 3$} $\to$ $100$ {\tiny $\pm 0$} & $99$ {\tiny $\pm 1$} $\to$ $100$ {\tiny $\pm 0$} & ${100}$ {\tiny $\pm 0$} $\to$ $100$ {\tiny $\pm 0$} \\
\texttt{scene-play-singletask-task4-v0} & $5$ {\tiny $\pm 1$} $\to$ $0$ {\tiny $\pm 0$} & $6$ {\tiny $\pm 3$} $\to$ $0$ {\tiny $\pm 0$} & $12$ {\tiny $\pm 6$} $\to$ $13$ {\tiny $\pm 8$} & ${20}$ {\tiny $\pm 9$} $\to$ $20$ {\tiny $\pm 12$} \\
\texttt{scene-play-singletask-task5-v0} & ${0}$ {\tiny $\pm 0$} $\to$ $0$ {\tiny $\pm 0$} & $0$ {\tiny $\pm 0$} $\to$ $0$ {\tiny $\pm 0$} & $0$ {\tiny $\pm 0$} $\to$ $0$ {\tiny $\pm 0$} & $0$ {\tiny $\pm 0$} $\to$ $0$ {\tiny $\pm 0$} \\
\midrule
\texttt{puzzle-3x3-play-singletask-task1-v0} & $90$ {\tiny $\pm 4$} $\to$ $97$ {\tiny $\pm 2$} & $90$ {\tiny $\pm 13$} $\to$ $100$ {\tiny $\pm 0$} & $91$ {\tiny $\pm 5$} $\to$ $100$ {\tiny $\pm 0$} & ${97}$ {\tiny $\pm 1$} $\to$ $100$ {\tiny $\pm 0$} \\
\texttt{puzzle-3x3-play-singletask-task2-v0} & ${16}$ {\tiny $\pm 5$} $\to$ $0$ {\tiny $\pm 0$} & $8$ {\tiny $\pm 4$} $\to$ $0$ {\tiny $\pm 0$} & $21$ {\tiny $\pm 2$} $\to$ $21$ {\tiny $\pm 25$} & ${36}$ {\tiny $\pm 1$} $\to$ $37$ {\tiny $\pm 1$} \\
\texttt{puzzle-3x3-play-singletask-task3-v0} & ${10}$ {\tiny $\pm 3$} $\to$ $0$ {\tiny $\pm 0$} & $8$ {\tiny $\pm 6$} $\to$ $0$ {\tiny $\pm 0$} & $12$ {\tiny $\pm 4$} $\to$ $12$ {\tiny $\pm 10$} & ${17}$ {\tiny $\pm 3$} $\to$ $16$ {\tiny $\pm 1$} \\
\texttt{puzzle-3x3-play-singletask-task4-v0 (*)} & ${16}$ {\tiny $\pm 5$} $\to$ $0$ {\tiny $\pm 0$} & $14$ {\tiny $\pm 6$} $\to$ $0$ {\tiny $\pm 0$} & $16$ {\tiny $\pm 5$} $\to$ $17$ {\tiny $\pm 13$} & ${21}$ {\tiny $\pm 5$} $\to$ $22$ {\tiny $\pm 5$} \\
\texttt{puzzle-3x3-play-singletask-task5-v0} & ${16}$ {\tiny $\pm 3$} $\to$ $0$ {\tiny $\pm 0$} & $20$ {\tiny $\pm 7$} $\to$ $0$ {\tiny $\pm 0$} & $23$ {\tiny $\pm 5$} $\to$ $23$ {\tiny $\pm 17$} & ${29}$ {\tiny $\pm 5$} $\to$ $30$ {\tiny $\pm 6$} \\
\midrule
\texttt{puzzle-4x4-play-singletask-task1-v0} & $34$ {\tiny $\pm 8$} $\to$ $100$ {\tiny $\pm 0$} & $20$ {\tiny $\pm 7$} $\to$ $100$ {\tiny $\pm 0$} & $44$ {\tiny $\pm 5$} $\to$ $100$ {\tiny $\pm 0$} & ${68}$ {\tiny $\pm 5$} $\to$ $100$ {\tiny $\pm 0$} \\
\texttt{puzzle-4x4-play-singletask-task2-v0} & ${16}$ {\tiny $\pm 5$} $\to$ $0$ {\tiny $\pm 0$} & $14$ {\tiny $\pm 3$} $\to$ $0$ {\tiny $\pm 0$} & $18$ {\tiny $\pm 2$} $\to$ $18$ {\tiny $\pm 14$} & ${24}$ {\tiny $\pm 2$} $\to$ $24$ {\tiny $\pm 3$} \\
\texttt{puzzle-4x4-play-singletask-task3-v0} & $18$ {\tiny $\pm 5$} $\to$ $100$ {\tiny $\pm 0$} & $16$ {\tiny $\pm 8$} $\to$ $100$ {\tiny $\pm 0$} & $31$ {\tiny $\pm 5$} $\to$ $100$ {\tiny $\pm 0$} & ${49}$ {\tiny $\pm 3$} $\to$ $100$ {\tiny $\pm 0$} \\
\texttt{puzzle-4x4-play-singletask-task4-v0 (*)} & $11$ {\tiny $\pm 3$} $\to$ $53$ {\tiny $\pm 50$} & $8$ {\tiny $\pm 1$} $\to$ $60$ {\tiny $\pm 71$} & $14$ {\tiny $\pm 2$} $\to$ $60$ {\tiny $\pm 47$} & ${20}$ {\tiny $\pm 3$} $\to$ $100$ {\tiny $\pm 0$} \\
\texttt{puzzle-4x4-play-singletask-task5-v0} & $7$ {\tiny $\pm 3$} $\to$ $0$ {\tiny $\pm 0$} & $4$ {\tiny $\pm 3$} $\to$ $0$ {\tiny $\pm 0$} & $9$ {\tiny $\pm 3$} $\to$ $9$ {\tiny $\pm 7$} & ${14}$ {\tiny $\pm 4$} $\to$ $13$ {\tiny $\pm 3$} \\
\midrule
\texttt{antmaze-umaze-v2} & ${96}$ {\tiny $\pm 2$} $\to$ $99$ {\tiny $\pm 1$} & $98$ {\tiny $\pm 4$} $\to$ $100$ {\tiny $\pm 1$} & $99$ {\tiny $\pm 2$} $\to$ $98$ {\tiny $\pm 0$} & ${100}$ {\tiny $\pm 0$} $\to$ $100$ {\tiny $\pm 0$} \\
\texttt{antmaze-umaze-diverse-v2} & ${89}$ {\tiny $\pm 5$} $\to$ $100$ {\tiny $\pm 0$} & $80$ {\tiny $\pm 8$} $\to$ $100$ {\tiny $\pm 0$} & $90$ {\tiny $\pm 4$} $\to$ $100$ {\tiny $\pm 0$} & ${96}$ {\tiny $\pm 0$} $\to$ $100$ {\tiny $\pm 0$} \\
\texttt{antmaze-medium-play-v2} & $78$ {\tiny $\pm 7$} $\to$ $98$ {\tiny $\pm 2$} & $74$ {\tiny $\pm 14$} $\to$ $98$ {\tiny $\pm 3$} & $78$ {\tiny $\pm 5$} $\to$ $98$ {\tiny $\pm 0$} & ${85}$ {\tiny $\pm 3$} $\to$ $100$ {\tiny $\pm 0$} \\
\texttt{antmaze-medium-diverse-v2} & $71$ {\tiny $\pm 13$} $\to$ $95$ {\tiny $\pm 1$} & $70$ {\tiny $\pm 8$} $\to$ $96$ {\tiny $\pm 1$} & $76$ {\tiny $\pm 4$} $\to$ $96$ {\tiny $\pm 0$} & ${85}$ {\tiny $\pm 3$} $\to$ $99$ {\tiny $\pm 2$} \\
\texttt{antmaze-large-play-v2} & ${84}$ {\tiny $\pm 7$} $\to$ $93$ {\tiny $\pm 6$} & $86$ {\tiny $\pm 7$} $\to$ $94$ {\tiny $\pm 8$} & $89$ {\tiny $\pm 4$} $\to$ $86$ {\tiny $\pm 0$} & ${91}$ {\tiny $\pm 1$} $\to$ $98$ {\tiny $\pm 0$} \\
\texttt{antmaze-large-diverse-v2} & ${83}$ {\tiny $\pm 4$} $\to$ $89$ {\tiny $\pm 2$} & $86$ {\tiny $\pm 3$} $\to$ $88$ {\tiny $\pm 3$} & $90$ {\tiny $\pm 2$} $\to$ $88$ {\tiny $\pm 3$} & ${95}$ {\tiny $\pm 1$} $\to$ $97$ {\tiny $\pm 1$} \\
\midrule
\texttt{pen-human-v1} & $53$ {\tiny $\pm 6$} $\to$ $134$ {\tiny $\pm 2$} & $50$ {\tiny $\pm 21$} $\to$ $135$ {\tiny $\pm 3$} & $58$ {\tiny $\pm 12$} $\to$ $135$ {\tiny $\pm 2$} & ${61}$ {\tiny $\pm 7$} $\to$ $135$ {\tiny $\pm 2$} \\
\texttt{pen-cloned-v1} & $74$ {\tiny $\pm 11$} $\to$ $142$ {\tiny $\pm 8$} & $81$ {\tiny $\pm 13$} $\to$ $143$ {\tiny $\pm 11$} & $82$ {\tiny $\pm 7$} $\to$ $134$ {\tiny $\pm 8$} & ${90}$ {\tiny $\pm 7$} $\to$ $144$ {\tiny $\pm 6$} \\
\texttt{pen-expert-v1} & $142$ {\tiny $\pm 6$} $\to$ $158$ {\tiny $\pm 3$} & $142$ {\tiny $\pm 9$} $\to$ $157$ {\tiny $\pm 5$} & $144$ {\tiny $\pm 3$} $\to$ $155$ {\tiny $\pm 2$} & ${147}$ {\tiny $\pm 1$} $\to$ $161$ {\tiny $\pm 1$} \\
\texttt{door-human-v1} & $0$ {\tiny $\pm 0$} $\to$ $83$ {\tiny $\pm 1$} & $0$ {\tiny $\pm 0$} $\to$ $83$ {\tiny $\pm 1$} & $4$ {\tiny $\pm 2$} $\to$ $83$ {\tiny $\pm 3$} & $6$ {\tiny $\pm 3$} $\to$ $82$ {\tiny $\pm 2$} \\
\texttt{door-cloned-v1} & $2$ {\tiny $\pm 1$} $\to$ $78$ {\tiny $\pm 20$} & $2$ {\tiny $\pm 2$} $\to$ $69$ {\tiny $\pm 27$} & $6$ {\tiny $\pm 2$} $\to$ $64$ {\tiny $\pm 23$} & ${6}$ {\tiny $\pm 5$} $\to$ $82$ {\tiny $\pm 22$} \\
\texttt{door-expert-v1} & ${104}$ {\tiny $\pm 1$} $\to$ $106$ {\tiny $\pm 0$} & $105$ {\tiny $\pm 2$} $\to$ $106$ {\tiny $\pm 0$} & $105$ {\tiny $\pm 1$} $\to$ $106$ {\tiny $\pm 0$} & ${105}$ {\tiny $\pm 0$} $\to$ $106$ {\tiny $\pm 0$} \\
\midrule
\texttt{hammer-human-v1} & $1$ {\tiny $\pm 1$} $\to$ $114$ {\tiny $\pm 5$} & $1$ {\tiny $\pm 2$} $\to$ $112$ {\tiny $\pm 6$} & $2$ {\tiny $\pm 1$} $\to$ $112$ {\tiny $\pm 6$} & ${4}$ {\tiny $\pm 1$} $\to$ $121$ {\tiny $\pm 8$} \\
\texttt{hammer-cloned-v1} & ${11}$ {\tiny $\pm 9$} $\to$ $129$ {\tiny $\pm 8$} & $3$ {\tiny $\pm 7$} $\to$ $126$ {\tiny $\pm 11$} & $12$ {\tiny $\pm 11$} $\to$ $126$ {\tiny $\pm 9$} & ${14}$ {\tiny $\pm 13$} $\to$ $130$ {\tiny $\pm 7$} \\
\texttt{hammer-expert-v1} & $125$ {\tiny $\pm 3$} $\to$ $132$ {\tiny $\pm 0$} & $126$ {\tiny $\pm 4$} $\to$ $132$ {\tiny $\pm 0$} & $127$ {\tiny $\pm 1$} $\to$ $132$ {\tiny $\pm 0$} & ${128}$ {\tiny $\pm 0$} $\to$ $133$ {\tiny $\pm 0$} \\
\texttt{relocate-human-v1} & ${0}$ {\tiny $\pm 0$} $\to$ $10$ {\tiny $\pm 8$} & $0$ {\tiny $\pm 0$} $\to$ $11$ {\tiny $\pm 11$} & $0$ {\tiny $\pm 0$} $\to$ $11$ {\tiny $\pm 10$} & $0$ {\tiny $\pm 0$} $\to$ $10$ {\tiny $\pm 6$} \\
\texttt{relocate-cloned-v1} & $-0$ {\tiny $\pm 0$} $\to$ $31$ {\tiny $\pm 27$} & $0$ {\tiny $\pm 0$} $\to$ $44$ {\tiny $\pm 35$} & $1$ {\tiny $\pm 0$} $\to$ $28$ {\tiny $\pm 33$} & $1$ {\tiny $\pm 0$} $\to$ $36$ {\tiny $\pm 19$} \\
\texttt{relocate-expert-v1} & ${107}$ {\tiny $\pm 1$} $\to$ $106$ {\tiny $\pm 1$} & $107$ {\tiny $\pm 2$} $\to$ $107$ {\tiny $\pm 2$} & $108$ {\tiny $\pm 2$} $\to$ $108$ {\tiny $\pm 3$} & ${109}$ {\tiny $\pm 0$} $\to$ $110$ {\tiny $\pm 0$} \\
\midrule
\texttt{visual-cube-single-play-singletask-task1-v0} & ${81}$ {\tiny $\pm 12$} & $84$ {\tiny $\pm 11$} & $81$ {\tiny $\pm 8$} & ${85}$ {\tiny $\pm 10$} \\
\texttt{visual-cube-double-play-singletask-task1-v0} & $21$ {\tiny $\pm 11$} & $22$ {\tiny $\pm 21$} & $20$ {\tiny $\pm 15$} & ${23}$ {\tiny $\pm 10$} \\
\texttt{visual-scene-play-singletask-task1-v0} & ${98}$ {\tiny $\pm 3$} & $100$ {\tiny $\pm 1$} & $99$ {\tiny $\pm 1$} & ${100}$ {\tiny $\pm 0$} \\
\texttt{visual-puzzle-3x3-play-singletask-task1-v0} & $94$ {\tiny $\pm 1$} & $90$ {\tiny $\pm 8$} & $90$ {\tiny $\pm 6$} & ${95}$ {\tiny $\pm 7$} \\
\texttt{visual-puzzle-4x4-play-singletask-task1-v0} & ${33}$ {\tiny $\pm 6$} & $26$ {\tiny $\pm 21$} & $24$ {\tiny $\pm 15$} & ${37}$ {\tiny $\pm 6$} \\
\bottomrule
\end{tabular}
\end{threeparttable}
}
\vspace{-10pt}
\end{table}

\end{document}